\documentclass[journal]{IEEEtran}

\usepackage{graphicx}
\usepackage[table]{xcolor}
\definecolor{rblue}{rgb}{0,0.5,1}
\usepackage[breaklinks,colorlinks]{hyperref}
\hypersetup{colorlinks, citecolor=rblue}

\usepackage{times}
\usepackage{epsfig}
\usepackage{graphicx}
\usepackage{amsmath}
\usepackage{amssymb}

\usepackage{lipsum}
\usepackage{stfloats}
\usepackage{multicol}
\usepackage{multirow}
\usepackage{bm}
\usepackage{etoolbox}

\usepackage{array}
\usepackage{tabulary}
\usepackage[table]{xcolor} 
\usepackage{paralist}
\usepackage{booktabs}
\usepackage{adjustbox}
\usepackage{xspace}
\makeatletter
\def\eg{\textit{e.g.}} 
\def\ie{\textit{i.e.}}

\makeatother

\usepackage[capitalize]{cleveref}
\crefname{section}{Sec.}{Secs.}
\Crefname{section}{Section}{Sections}
\Crefname{table}{Table}{Tables}
\crefname{table}{TABLE}{Tabs.}
\definecolor{rblue}{rgb}{0,0.5,1}

\newcommand{\PAR}[1]{\noindent{\bf #1}}

\begin{document}

\title{Computational Imaging for Machine Perception: Transferring Semantic Segmentation\\beyond Aberrations}

\author{Qi Jiang\IEEEauthorrefmark{1}, Hao Shi\IEEEauthorrefmark{1}, Shaohua Gao, Jiaming Zhang, Kailun Yang\IEEEauthorrefmark{2}, Lei Sun, Huajian Ni, and Kaiwei Wang\IEEEauthorrefmark{2}%
\thanks{This work was supported in part by the National Natural Science Foundation of China (NSFC) under Grant No. 12174341 and in part by Hangzhou SurImage Technology Company Ltd.}%
\thanks{Q. Jiang, H. Shi, S. Gao, L. Sun, and K. Wang are with the State Key Laboratory of Extreme Photonics and Instrumentation and the National Engineering Research Center of Optical Instrumentation, Zhejiang University, Hangzhou 310027, China.}%
\thanks{J. Zhang is with the Institute for Anthropomatics and Robotics, Karlsruhe Institute of Technology, Karlsruhe 76131, Germany.}
\thanks{K. Yang is with the School of Robotics and the National Engineering Research Center of Robot Visual Perception and Control Technology, Hunan University, Changsha 410082, China.}%
\thanks{H. Shi and H. Ni are with Shanghai SUPREMIND Technology Company Ltd, Shanghai 201210, China.}%
\thanks{\IEEEauthorrefmark{1}Equal contribution.}%
\thanks{\IEEEauthorrefmark{2}Corresponding authors: Kaiwei Wang and Kailun Yang. (E-mail: wangkaiwei@zju.edu.cn, kailun.yang@hnu.edu.cn.)}%
}

\markboth{IEEE Transactions on Computational Imaging, March~2024}%
{Jiang~\MakeLowercase{\textit{et al.}}: CIADA}

\maketitle

\begin{abstract}
Semantic scene understanding with Minimalist Optical Systems (MOS) in mobile and wearable applications remains a challenge due to the corrupted imaging quality induced by optical aberrations. However, previous works only focus on improving the subjective imaging quality through the Computational Imaging (CI) technique, ignoring the feasibility of advancing semantic segmentation. In this paper, we pioneer the investigation of Semantic Segmentation under Optical Aberrations (SSOA) with MOS. To benchmark SSOA, we construct \emph{Virtual Prototype Lens (VPL)} groups through optical simulation, generating \emph{Cityscapes-ab} and \emph{KITTI-360-ab} datasets under different behaviors and levels of aberrations. We look into SSOA via an unsupervised domain adaptation perspective to address the scarcity of labeled aberration data in real-world scenarios. Further, we propose \emph{Computational Imaging Assisted Domain Adaptation (CIADA)} to leverage prior knowledge of CI for robust performance in SSOA. Based on our benchmark, we conduct experiments on the robustness of classical segmenters against aberrations. In addition, extensive evaluations of possible solutions to SSOA reveal that CIADA achieves superior performance under all aberration distributions, bridging the gap between computational imaging and downstream applications for MOS. The project page is at \url{https://github.com/zju-jiangqi/CIADA}.
\end{abstract}

\begin{IEEEkeywords}
Computational imaging, domain adaptation, semantic segmentation, optical aberrations, scene parsing.
\end{IEEEkeywords}

\IEEEpeerreviewmaketitle

\section{Introduction}
\begin{figure}[!t]
  \centering
  \includegraphics[width=1.0\linewidth]{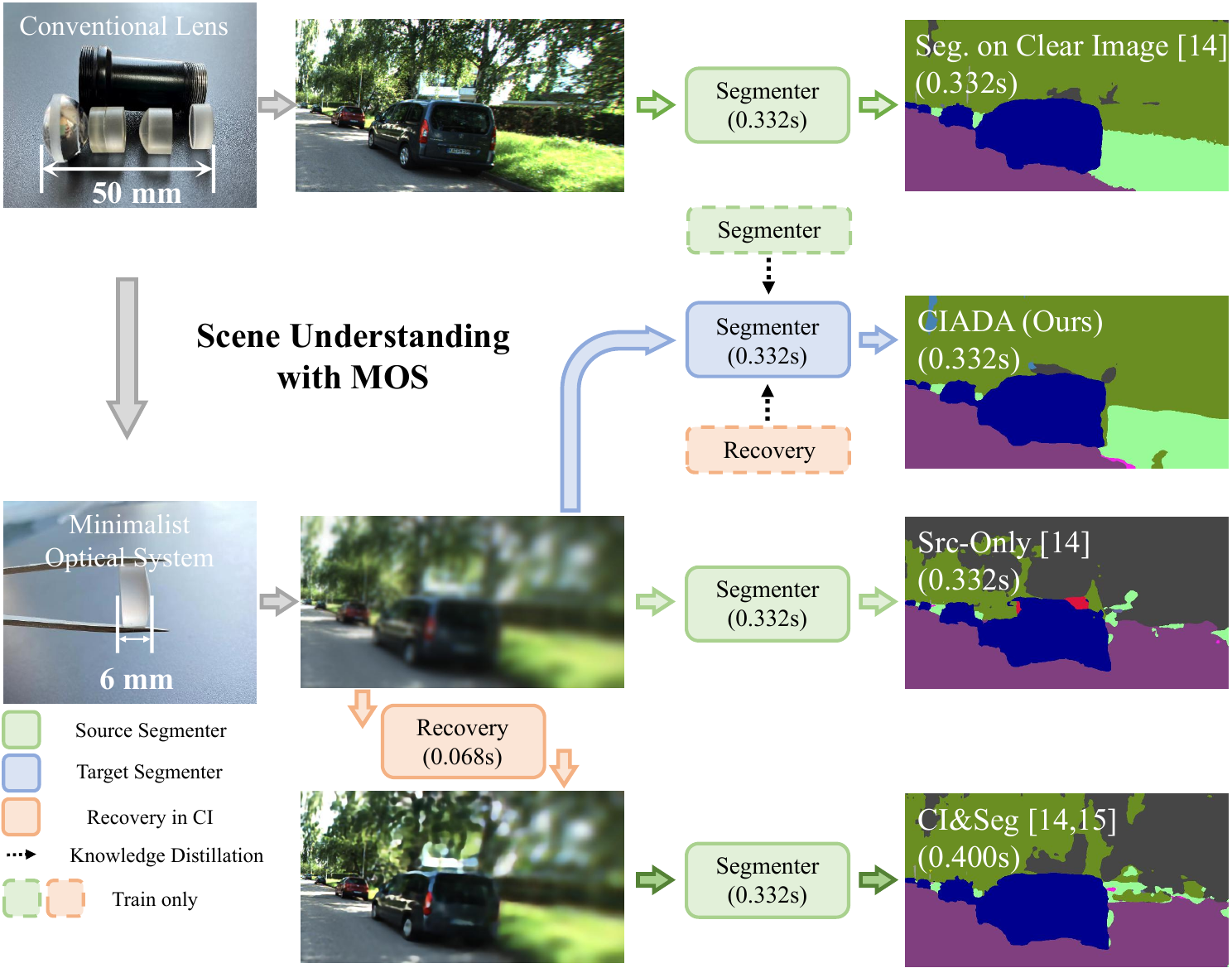}
  \caption{We pioneer scene understanding with Minimalist Optical Systems (MOS). The proposed \emph{Computational Imaging Assisted Domain Adaptation (CIADA)} can achieve more accurate segmentation on the aberration image of MOS, compared to other possible solutions, \eg~Src-Only (SegFormer~\cite{xie2021segformer}) and CI\&Seg (applying NAFNet~\cite{chen2022simple} before SegFormer~\cite{xie2021segformer}), without the extra computational overhead. The result is comparable to conventional lenses with sophisticated lens designs. The inference time of each model and the total inference time of each possible solution are shown in the ``()''.} 
  \label{fig:intro}

\end{figure}

\IEEEPARstart{S}{emantic} scene understanding attracts considerable research interest for its potential applications in autonomous driving~\cite{cordts2016cityscapes}, augmented reality~\cite{zhang2022trans4trans}, and robot navigation~\cite{le2022bayesian}.
As the demand for thin, portable imaging systems grows stronger in mobile and wearable applications, \eg, navigation aids for the blind~\cite{yang2018unifying,liu2023open}, augmented reality devices~\cite{cui2020development}, surveillance drones~\cite{sun2021aerial,wang2022high}, mobile phones~\cite{zhang2023large}, and search and rescue robots~\cite{uddin2016search,deng2020semantic}, semantic perception with Minimalist Optical Systems (MOS)~\cite{hua2022ultra,liu2022computational}, which can image over a large Field of View (FoV) with only a few optical elements (shown in Fig.~\ref{fig:intro}), has come to the research forefront.

\begin{figure*}[!t]
  \centering
  \includegraphics[width=1.0\linewidth]{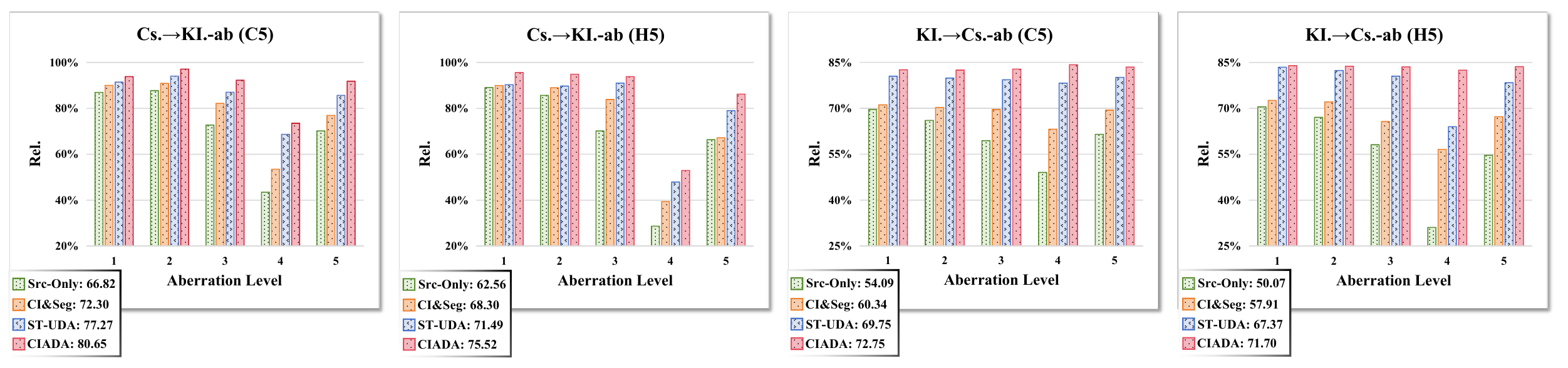}
  \caption{Relative performance (Rel., the normalized mIoU after being divided by that of the score of the Oracle) of possible solutions to Semantic Segmentation under Optical Aberrations (SSOA). Following~\cite{michaelis2019benchmarking,hummer2023vltseg}, we also apply the robustness metric of relative performance under corruption (rPC) over the four aberration levels for a comprehensive evaluation, which is shown in the bottom left corner of the figure ($\%$). The aberration level is the defined incremental level for different severity of aberrations, which will be depicted in Sec.~\ref{sec:benchmark}. The relative results reveal that CIADA outperforms other solutions at all levels.}
  \label{fig:solution}
\end{figure*}

The degraded imaging quality induced by optical aberrations is an inevitable problem of MOS, where Computational Imaging (CI) technologies appear as a preferred solution.
In recent CI pipelines~\cite{dun2020learned,Peng2019LearnedLF}, the captured images are recovered through learning-based models for comparable visual results to conventional lenses.
However, a recovered image seems limited to high-level tasks:
(1)~Previous methods only recover images optimized for human observation, ignoring the \textbf{\emph{data distribution for downstream tasks}}~\cite{diamond2021dirty}, which may lead to unpleasant performance~\cite{diamond2017dirty,sakaridis2018semantic}. 
(2)~\textbf{\emph{The extra computation overhead}} of recovery also hinders their implementations in real applications. 
Considering that most CI solutions can hardly solve the weaknesses of MOS in scene understanding~\cite{pei2019effects,vidalmata2020bridging}, we explore the direct segmentation task on images of MOS, \ie~\textbf{{Semantic Segmentation under Optical Aberrations (SSOA)}}, which has been barely investigated.

One major challenge lies in the scarcity of labeled training data, corrupted by aberrations, for yielding robust segmentation models~\cite{cordts2016cityscapes,liao2022kitti}.
Yet, it is cost-intensive to acquire large-scale pixel-wise labeled data, in particular for aberration images which are harder to annotate.
While imaging simulation can generate target images from existing datasets, the synthetic-to-real gap remains a potential problem.
Meanwhile, the recent progress in Unsupervised Domain Adaptation (UDA) has brought promising solutions~\cite{hoyer2022daformer,luo2019clan,zou2019crst}, which offers a novel perspective to supplement the coverage of training data in unseen scenarios.
In this paper, we look at SSOA through the lens of UDA by distilling knowledge from label-rich clean data to label-scarce aberration data.

To this intent, we formalize the task of UDA for SSOA in a comprehensive benchmark 
through wave-based imaging simulation~\cite{mahajan1994zernike}.
We summarize aberration behaviors of MOS: Common Simple Lens (CSL, \ie~spatial-variant degradation) and Hybrid Refractive Diffractive Lens (HRDL, \ie~spatial-uniform degradation), and construct \emph{Virtual Prototype Lens (VPL)} groups to produce simulated imaging results of different MOS.
VPL groups contain randomly generated aberration distributions of four levels for CSL and HRDL, respectively.
Specifically, we create \emph{Cityscapes-ab} and \emph{KITTI-360-ab} to benchmark semantic segmentation under various optical aberrations. 
The task of our UDA setting is to train the segmenter to bridge the aberration-induced domain gap.
In this case, \emph{the situations of real and synthetic data are equivalent}, as both processes can be interpreted as domain adaptation from labeled clear images to unlabeled aberration images. 

Then, we propose the \emph{Computational Imaging Assisted Domain Adaptation (CIADA)} as a solution to domain adaptive SSOA.
CIADA is built on a self-training UDA (ST-UDA) baseline DAFormer~\cite{hoyer2022daformer}, where the transformer-based encoder reveals stronger robustness to image corruptions~\cite{xie2021segformer}. 
We put forward the \emph{Bidirectional Teacher (BT)} to store the implicit image degradation knowledge from CI through image degradation and recovery, which is applied to generate augmented images for the source domain and pseudo-ground-truth images for the target domain during UDA training. 
Further, an auxiliary decoder for image recovery with correlation-based distillation is designed to distill the CI knowledge to the target segmenter via multi-task learning.
Compared to the ST-UDA baseline, CIADA transfers the knowledge of aberration-induced blur from CI to the target domain for SSOA, without additional computational overhead during inference.

Based on the benchmark, we conduct comprehensive evaluations on the influence of aberrations on state-of-the-art segmenters, demonstrating that aberrations bring considerable damage to the segmentation performance. 
Additionally, extensive experiments on possible solutions to SSOA, \eg~Src-Only (model trained on source clear images), CI\&Seg (computational imaging before Src-Only model) and proposed CIADA, as shown in Fig.~\ref{fig:intro}, show that CIADA is a more competitive solution. Taking the Oracle (model trained on target aberration images) as the upper limit, we also show the relative performance and the results under the robustness metric of relative performance under corruption (rPC)~\cite{michaelis2019benchmarking,hummer2023vltseg} over the four aberration levels in Fig.~\ref{fig:solution}, where CIADA effectively mitigates the clear-to-aberration domain gap and reveals superiority in all settings. Specifically, CIADA outperforms the baseline ST-UDA method~\cite{hoyer2022daformer} by $3.29/4.02$ in mIoU on {Cityscapes~\cite{cordts2016cityscapes}}${\to}${KITTI-360-ab} and $2.28/3.48$ in mIoU on {KITTI-360~\cite{liao2022kitti}}${\to}${Cityscapes-ab}.
CIADA is a successful attempt to combine advanced CI techniques with unsupervised domain adaptive SSOA, presenting a novel insight and paradigm for applying MOS in semantic scene understanding.  

In summary, \emph{it is the first work to address semantic segmentation under severe optical aberrations of MOS}, to the best of our knowledge.
The conventional computational imaging for human perception is extended to \textbf{computational imaging for machine perception}.
We deliver four main contributions: 
\begin{compactitem}
    \item We explore the applications of minimalist optical systems in semantic scene understanding, \ie, semantic segmentation beyond optical aberrations, from an unsupervised domain adaptation perspective.
    \item We benchmark the task through \emph{Virtual Prototype Lens (VPL)} groups, which can translate existing datasets into corrupted images under various levels of aberrations. We provide \emph{Cityscapses-ab} and \emph{KITTI-360-ab} to foster semantic perception under aberrations. 
    \item We propose \emph{Computational Imaging Assisted Domain Adaptation}, coined \emph{CIADA}, a UDA framework distilling the knowledge of aberration-induced blur from computational imaging to segmentation under aberrations.
    \item We conduct comprehensive evaluations on classical segmenters and possible solutions to SSOA, where UDA is verified effective for SSOA by experiments and CIADA provides a superior solution. 
\end{compactitem}


\section{Related Work}
\label{sec:related_work}
\subsection{Computational Imaging for Minimalist Optical Systems} 
To make a sweet spot between the trade-off of high-quality imaging and lightweight lenses, Computational Imaging (CI)~\cite{barbastathis2019use} technologies are often applied to the compensation for the optical aberrations of MOS.
Early work centers around simple lenses composed of a few spherical- or aspherical lenses, where deconvolution~\cite{schuler2011non,wu2020non} is often used for image recovery.
Owing to powerful learning-based models~\cite{chen2022simple,zamir2022restormer}, plentiful CI pipelines~\cite{chen2021optical,10021856,li2021universal} apply image recovery networks.
Meanwhile, advanced optical design techniques, \eg~Diffractive Optical Elements (DOE)~\cite{liu2022computational,peng2016diffractive}, Fresnel lens~\cite{Peng2019LearnedLF}, and meta-lens~\cite{hua2022ultra}, equip MOS with capacities of imaging over a large Field of View (FoV) with a single optical element.
However, the potential applications of MOS in semantic scene understanding are rarely explored.
Almost all CI pipelines only focus on clear images optimized for human observation, while leading to corrupted data distribution for segmenters~\cite{pei2019effects}.
The worse performance~\cite{vidalmata2020bridging} and additional computational overhead make CI hardly adaptable to existing semantic segmentation frameworks~\cite{xie2021segformer, chen2018encoder,guo2022segnext,long2015fully}.
To this end, we investigate performing semantic segmentation directly on aberration images, \ie~SSOA.
The novel task delivers a new insight into computational imaging, which targets machine perception rather than human perception. 

\begin{figure}[!t]
  \centering
  \includegraphics[width=0.8\linewidth]{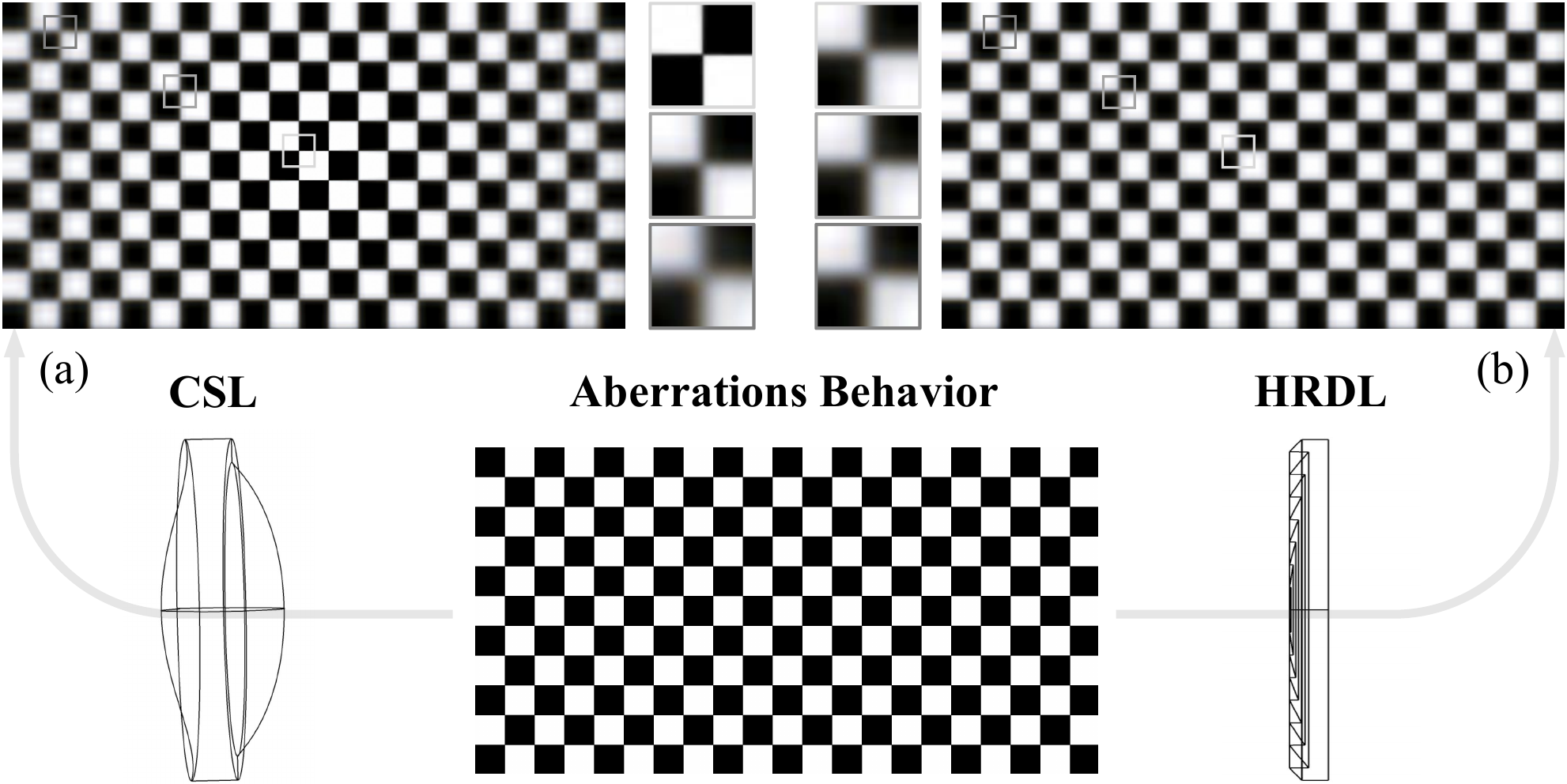}
  \caption{We summarize two aberrations behaviors for MOS: (a) Common Simple Lens (CSL), revealing spatial-variant degradation. (b) Hybrid Refractive Diffractive Lens (HRDL), revealing spatial-uniform degradation.}
  \label{fig:behavior}
\end{figure}

\subsection{Semantic Segmentation under Image Corruptions}
To enhance semantic understanding in real-world conditions, different degradation scenarios are considered~\cite{guo2019degraded,kamann2020increasing_painting}.
Nighttime segmentation aims to produce robust parsing under low illuminations~\cite{romera2019bridging,liu2023improving_nighttime}, whereas various adverse weather situations are covered in~\cite{zhang2022trans4trans,sakaridis2021acdc}.
The WildDash benchmark~\cite{zendel2018wilddash} is raised to improve hazard awareness.
In~\cite{kamann2020benchmarking}, a benchmark is created with a wide spectrum of degradation conditions spanning blur, noise,
and digital corruptions, while the segmentation robustness against these corruption variants is largely enhanced via recent transformer models~\cite{xie2021segformer,zhou2022fan}.
Yet, the mentioned aberration-based PSF blur is different from that of MOS, which hardly influences the performance of contemporary models. 
In addition, the OpticsBench benchmark~\cite{muller2023classification} is proposed for investigating robustness to optical blur effects, but only the effects of decomposed single-class aberrations are considered, which do not exist in the aberration distributions of MOS.
In this work, we delve deeper into aberration-induced degradation and offer a comprehensive benchmark, which covers image degradation under diverse aberration distributions of MOS.
Unlike previous works, we further view semantic segmentation under aberrations via an unsupervised domain adaptation perspective to yield robust segmentation against unseen aberration images.

\subsection{Unsupervised Domain Adaptive Segmentation}
Domain adaptation has been frequently investigated to improve segmentation model generalization to new, unseen scenarios.
Two main paradigms of domain adaptation include self-training methods~\cite{zou2019crst,lian2019constructing_self_motivated,zhang2017curriculum_domain_adaptation,9785619} and adversarial methods~\cite{luo2019clan,chang2019all_about_structure,tsai2018adaptsegnet,9889741}.
The self-training solution gradually optimizes the model via generated pseudo labels.
The adversarial strategy builds on the idea of GANs~\cite{goodfellow2020gan} to enable image translation or enforce agreement via layout matching~\cite{huang2020contextual_relation_consistent,li2020content_consistent} or feature alignment~\cite{luo2019clan,wang2020differential_treatment}.
Recent adaptation techniques include 
uncertainty reduction~\cite{zheng2021rectifying_pseudo_label},
Fourier transforms~\cite{yang2020fda},
cycle association~\cite{kang2020pixel_cycle_association},
entropy minimization~\cite{vu2019advent},
prototypical regularization~\cite{zhang2021proda},
contrastive learning~\cite{xie2022sepico},
cross-domain mixed sampling~\cite{tranheden2021dacs,9889681}, 
and transformers~\cite{hoyer2022daformer,hoyer2022hrda}.
Differing from these works, we propose a \emph{Computational Imaging Assisted Domain Adaptation (CIADA)} framework for robust segmentation under optical aberrations, exploring the potential of domain adaptive SSOA with an auxiliary CI pipeline.

\section{Benchmark for Semantic Segmentation under Optical Aberrations}
\label{sec:camera}
In this section, we set up the benchmark for SSOA by constructing \emph{Virtual Prototype Lens (VPL)}. 
We first analyze the behavior of aberrations of MOS in Sec.~\ref{sec:behavior}. Then, the wave-based imaging simulation is detailed in Sec.~\ref{sec:simulation}, based on which we construct the \emph{Virtual Prototype Lens(VPL)} for the generation of images under various aberration distributions in Sec.~\ref{sec:vpl}.
Finally, we elaborate on the benchmark and corresponding settings in Sec.~\ref{sec:benchmark}. 
VPL is a landmark engine for aberration data generation, which enables comprehensive evaluations of potential solutions to SSOA. 

\begin{figure}[!t]
  \centering
  \includegraphics[width=1\linewidth]{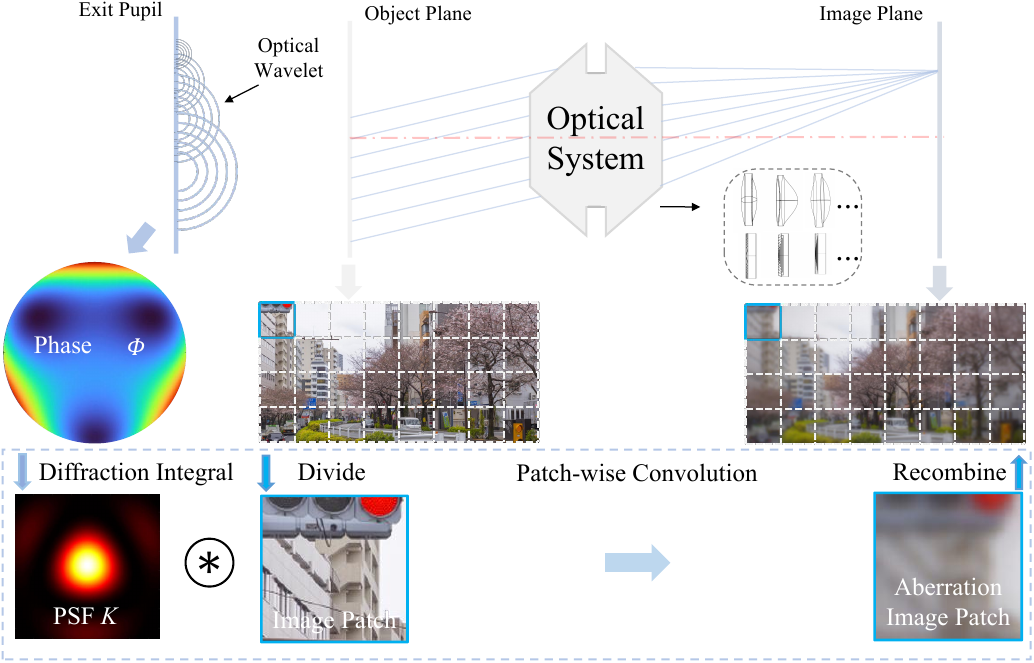}
  \caption{The simulation pipeline for constructing virtual prototype lens. The optical system is considered a black box with phase function $\Phi$. We can simulate the imaging result of any MOS through the pipeline. }
  \label{fig:pipeline}
\end{figure}

\begin{figure*}[!t]
  \centering
  \includegraphics[width=1.0\linewidth]{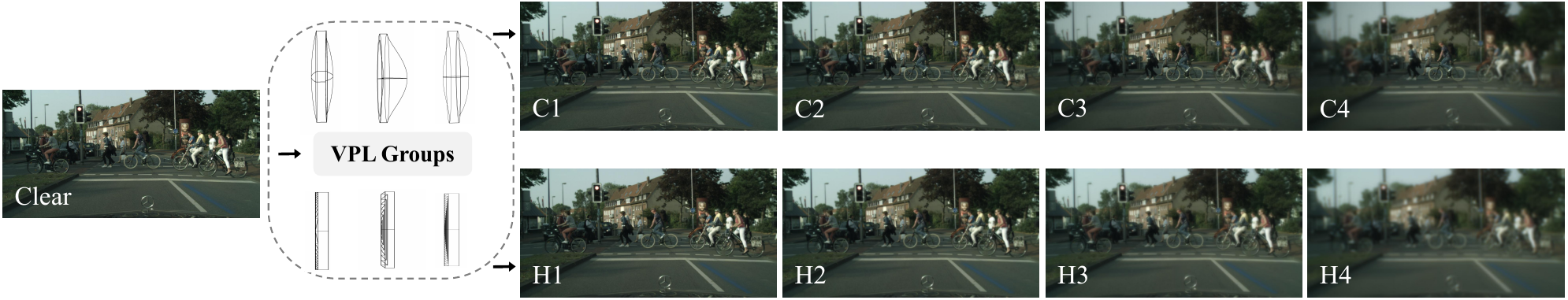}
  \caption{A novel benchmark for \emph{Semantic Segmentation under Optical Aberrations (SSOA)}. We construct \emph{Virtual Prototype Lens (VPL)} groups with different behaviors and levels of aberrations to map clear images into simulated imaging results. C1-C4: aberration levels for CSL, H1-H4: aberration levels for HRDL. A hybrid set of C1/H1 to C4/H4 composes the level C5/H5, for comprehensive training and evaluation of SSOA solutions. }
  \label{fig:benchmark}
\end{figure*}

\subsection{Analysis of Aberrations Behavior}
\label{sec:behavior}

Although aberrations of various optical systems are distinctly different, a summary of the commonalities and characteristics is necessary for a comprehensive benchmark for SSOA.
For most MOS, the aberration distribution reveals two behaviors: \emph{spatial-variant degradation}~\cite{schuler2011non,li2021universal} and \emph{spatial-uniform degradation}~\cite{liu2022computational,Peng2019LearnedLF}, which are relevant to the imaging principle of the optical components. To be specific, the \emph{Common Simple Lens (CSL)} induces spatial-variant degradation, whereas the \emph{Hybrid Refractive Diffractive Lens (HRDL)} induces spatial-uniform degradation, as shown in Fig.~\ref{fig:behavior}.

\PAR{Common simple lens.}
CSL often consists of optical elements with continuous surfaces~\cite{chen2021optical}, \ie~spherical and aspherical surfaces. Due to refraction effects, the optical path varies with different thicknesses of the center and the edge FoVs, which ultimately leads to imaging results with spatial-variant degradation over different FoVs.
As illustrated in Fig.~\ref{fig:behavior}(a), the image patch in the center of the FoV, which is associated with the paraxial region, is clear, while it degrades gradually with the increasing viewing angle.   

\PAR{Hybrid refractive diffractive lens.}
With the booming processing technology of micro- and nano-precision optical components, Fresnel and DOE are applied in the design of  MOS~\cite{Peng2019LearnedLF}.
The imaging principle of this type of MOS includes refraction and diffraction, called HRDL.
HRDL enables a more uniform thickness distribution of the optical elements, avoiding conspicuous differences in their optical path.
In this sense, the aberrations of HRDL induce spatial-uniform degradation, \ie~the image patches at the center and edge viewing angles degrade similarly, shown in Fig.~\ref{fig:behavior}(b).

For image recovery in CI pipelines, the solutions to spatial-variant and spatial-uniform degradation are often different~\cite{Peng2019LearnedLF,chen2021optical}.
Consequently, our established benchmark for SSOA is also divided accordingly based on the aberrations behaviors of CSL and HRDL. 
\begin{figure}[!t]
  \centering
  \includegraphics[width=1.0\linewidth]{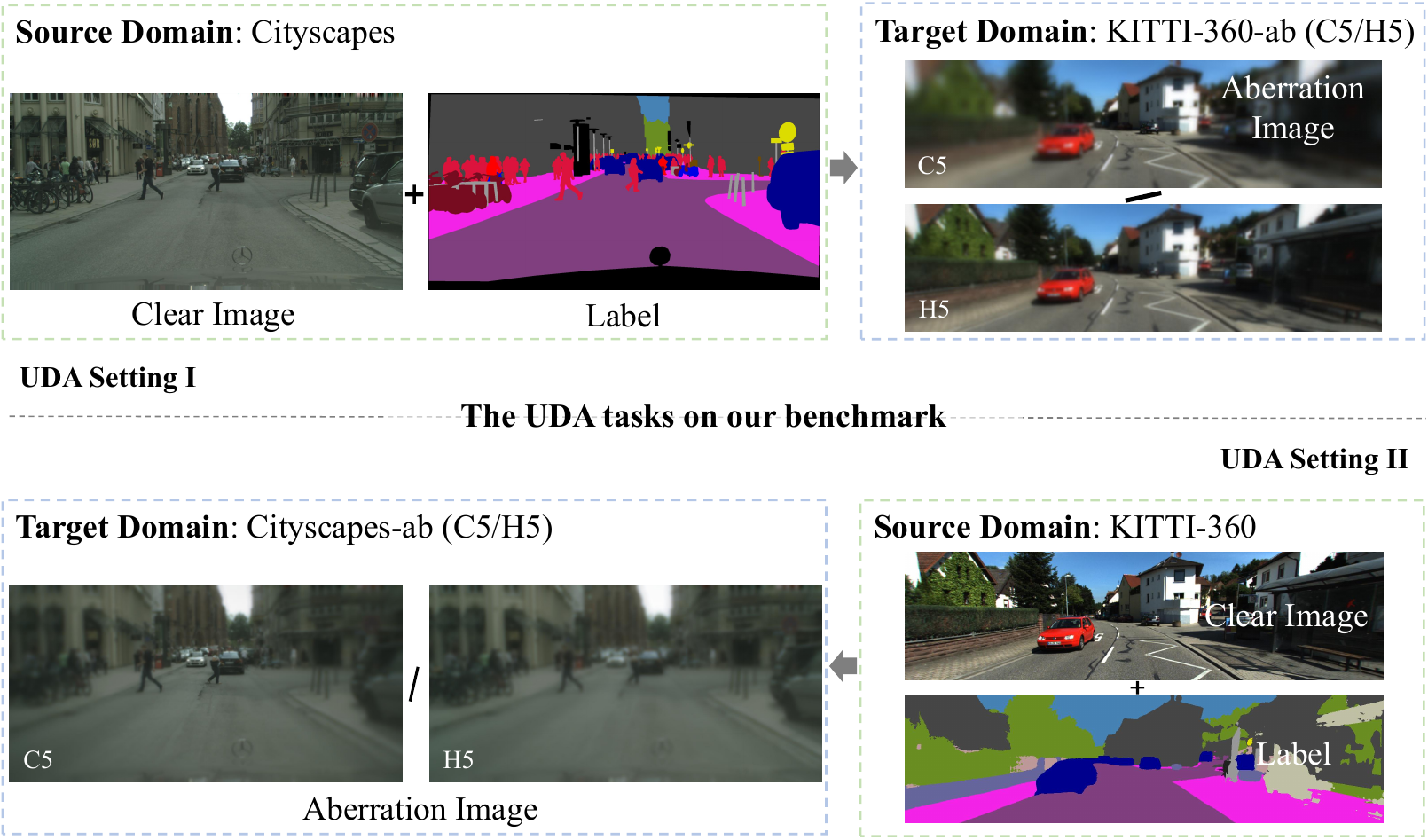}
  \caption{Illustration of the UDA tasks on our benchmark. We establish two settings to investigate SSOA without access to the labels of images with aberration.}
  \label{fig:task}
\end{figure}

\subsection{Wave-based Imaging Simulation}
\label{sec:simulation}

In learning-based CI pipelines, simulation~\cite{chen2021optical} or live-shooting~\cite{Peng2019LearnedLF} is applied to generate data pairs for network training.
When it comes to SSOA, imaging simulation appears as a more flexible solution to the acquisition of large-scale labeled data under various aberration distributions. 

Considering the difficulty of ray-tracing between distinct components of blind MOS, we view the imaging process in an abstract way.
Applicable to most MOS, the modulation of the system on the incident wave phase is reflected in the phase function on its exit pupil plane.
In other words, the specific structure of MOS can be abstracted into a black box, while the imaging process can be simulated only by modeling the phase function on exit pupil, \ie~wave-based imaging simulation.

To be specific, we apply Zernike polynomials~\cite{mahajan1994zernike} to describe the phase function of optical wavefronts $\Phi(x',y',\theta,\lambda)$ on exit pupil mathematically:
\begin{equation}
\label{eq:zernike}
\Phi(x',y',\theta,\lambda) = \sum_{n,m} {C^m_n}(\theta,\lambda){Z^m_n}(x',y'),
\end{equation}
where $C(\theta,\lambda)$ denotes Zernike coefficients under FoV $\theta$ and wavelength $\lambda$ and $Z$ refers to polynomials of the coordinate $(x',y')$ on exit pupil.
The combination of different $m$ and $n$ represents different orders.
For the specific expressions of each item, please refer to~\cite{mahajan1994zernike}. 

$\Phi(x',y',\theta,\lambda)$ represents the modulation of any MOS on the incident light wave of $\theta$ and $\lambda$. It can be converted to Point Spread Function (PSF) $K(x,y,\theta,\lambda)$ on the image plane through scalar diffraction integral~\cite{huggins2007introduction} (detailed in the Appendix).
The impacts of off-axis aberration and chromatic aberration are also considered in the pipeline via phase functions under different $\theta$ and $\lambda$.

In this way, we can simulate the imaging result of any MOS through patch-wise convolution~\cite{chen2021optical} between the clear image and PSFs of different FoVs and wavelengths. The simulation pipeline is shown in Fig.~\ref{fig:pipeline}.

\subsection{Virtual Prototype Lens}
\label{sec:vpl}
We construct \emph{Virtual Prototype Lens (VPL)} groups based on the proposed simulation pipeline to transform clear images to corrupted ones under various aberration distributions, as shown in Fig~\ref{fig:benchmark}.
To further investigate semantic segmentation under different severity of aberrations, we set up four aberration levels for CSL and HRDL respectively. The level is defined according to the shape (distribution of Zernike coefficients) and the kernel size (distribution of radius of spot diagram) of PSFs.
Concretely, C1 to C4 are incremental levels for CSL, whereas an additional level C5 provides a hybrid set of C1 to C4 for comprehensive evaluation. 
Similarly, H1 to H5 are set for HRDL. 

In detail, the curves of different Zernike coefficients of real MOS samples under different severity of aberrations, as a function of FoVs, are statistically analyzed.
Then, we set up the range and curve trend of Zernike coefficients for each level based on the statistical analysis.
In the same way, the random range of each aberration level for the radius of the spot diagram is also established.
For detailed statistics of real MOS samples, please refer to the Appendix. 

Supplied with the random range, we generate random distributions of Zernike coefficients over $128$ normalized FoVs to describe the phase function in Eq.~\ref{eq:zernike}, \ie~different VPLs of CSL and HRDL.
In the case of CSL (the same is true for HRDL), we produce five VPL samples for C1-C4 respectively. Then, we fine-tune the ranges of four levels and generate five new samples for each level to produce C5, where the samples in C5 are not duplicated with those in C1-C4, serving as an additional hybrid level. 
Hence, in Fig.~\ref{fig:benchmark}, we generate simulated results under different behaviors and levels of aberrations for any clear image datasets via VPL samples. 

\subsection{Benchmark Settings}
\label{sec:benchmark}

\begin{figure*}[!t]
  \centering
  \includegraphics[width=1.0\linewidth]{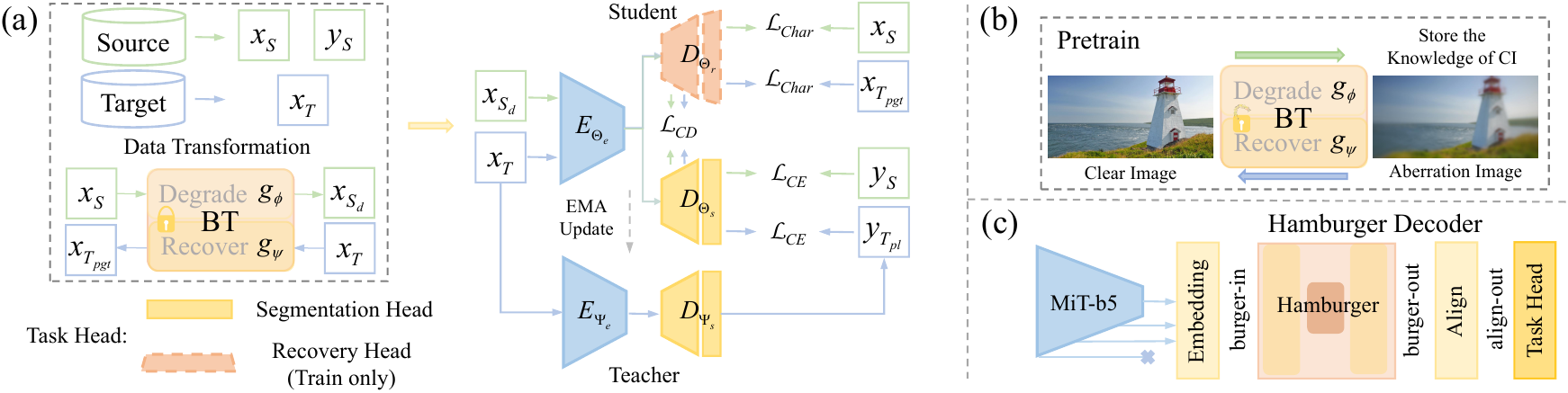}
  \caption{Our solution to unsupervised domain adaptive semantic segmentation under aberrations. (a) Overview of the CIADA architecture. We apply data transformation and auxiliary recovery tasks to distill the knowledge of CI. (b) Bidirectional Teacher (BT) for storing the knowledge of CI. (c) The network architecture of CIADA. Following~\cite{guo2022segnext}, our decoder only aggregates the last three stages of features from the encoder. The features from `burger-in', `burger-out', and `align-out' can be used to calculate the CD Loss. }
  \label{fig:ciada}
\end{figure*}

Cityscapes~\cite{cordts2016cityscapes} and KITTI-360~\cite{liao2022kitti} are two semantic segmentation datasets with a large number of clear images and fine-annotated labels, where the categories and annotation methods are also consistent.
In our work, we create \emph{Cityscapes-ab} and \emph{KITTI-360-ab} through our established VPL groups for benchmarking SSOA, which include labeled images under various aberration distributions. 

The clear images from the original datasets are modulated by VPL samples of C1-C5 and H1-H5, while the labels remain unchanged. The evaluation of semantic segmentation under different severity of aberrations is performed on the validation set of C1-C5 and H1-H5, where C5 and H5 may present comprehensive results.
Considering the synthetic-to-real gap induced by simulation and adaptability to real MOS, the task of our benchmark follows~\emph{unsupervised domain adaptive semantic segmentation}.
Specifically, as is shown in Fig.~\ref{fig:task}, UDA is performed on {Cityscapes}${\to}${KITTI-360-ab} (C5/R5) and {KITTI-360}${\to}${Cityscapes-ab} (C5/R5), which reflects real-world application scenarios where one cannot access the labels of corrupted images in an unseen domain taken by an unknown MOS.
In this way, the performance of our task is a reflection of how well it works on real MOS. 

In summary, our benchmark includes datasets under five levels of optical aberrations of two behaviors respectively.
The possible solutions to SSOA will be evaluated for the robustness and effectiveness of MOS's potential applications in semantic scene understanding on the benchmark. 

\section{Computational Imaging Assisted Domain Adaptation}
\label{sec:ciada}

In this section, we propose a UDA framework that transfers the knowledge of semantic segmentation under clear images and CI to SSOA, \ie~\emph{Computational Imaging Assisted Domain Adaptation (CIADA)}.
The basic setting of ST-UDA is first established as the baseline of CIADA in Sec.~\ref{sec:studa}.
Then, we detail the architecture and training pipeline of CIADA in Sec.~\ref{sec:arch} and Sec.~\ref{sec:training}, respectively.
As Fig.~\ref{fig:ciada} illustrates, CIADA explores coupling CI techniques with domain adaptive semantic segmentation via knowledge distillation.

\subsection{Self-Training based UDA}
\label{sec:studa}

With the goal of attaining stable and high performance of UDA~\cite{hoyer2022daformer}, self-training (ST) approaches~\cite{zou2019crst,lian2019constructing_self_motivated,zhang2017curriculum_domain_adaptation, araslanov2021self_supervised_augmentation_consistency,huo2022domain_agnostic_prior,lai2022decouplenet,li2022class_balanced_pixel_level_self_labeling} have achieved spectacular progress, where the model is gradually optimized via generated pseudo-labels.
For our specific UDA task towards SSOA, we build up our baseline solution as a basic ST-UDA framework.

The encoder $E_{\Theta_e}$ and segmentation decoder $D_{\Theta_s}$ are trained with labeled source data of clear images $\mathcal{D}_{S}=\{{x_S}^{(i)}, {y_S}^{(i)}\}_{i=1}^{N_S}$ and unlabeled target data of images with aberrations $\mathcal{D}_{T}=\{{x_T}^{(i)}\}_{i=1}^{N_T}$.
A teacher network with encoder $E_{\Psi_e}$ and decoder $D_{\Psi_s}$ is applied to generate pseudo-labels ${y_T}_{pl}$ for target data. Our baseline follows the UDA training pipeline of~\cite{hoyer2022daformer}, noted as `ST-UDA' in this paper.

\subsection{CIADA Architecture}
\label{sec:arch}
For ST-UDA, the image degradation induced by aberrations of MOS leads to difficulty in the optimization of pseudo-labels.
In other words, the domain gap is large, whereas the source knowledge of the clear domain seems to be insufficient. 
Meanwhile, computational imaging pipeline~\cite{dun2020learned,Peng2019LearnedLF}, where the image degradation of MOS is analyzed and processed by computational methods, delivers valuable guidance for understanding the aberration images.
Benefiting from the multi-task learning framework of semantic segmentation and low-level vision~\cite{wang2020dual}, we explore to promote the ST-UDA baseline via an auxiliary CI task.

To this intent, we propose CIADA, pioneering to leverage the additional knowledge from image recovery of CI for assisting ST-UDA. The architecture of CIADA is shown in Fig.~\ref{fig:ciada}(a). The auxiliary task, namely self-training-based image recovery, is established for knowledge distillation. 

\PAR{Bidirectional teacher of computational imaging.} 
To inform the segmenter of the instructive knowledge of image degradation, a teacher model is constructed based on the CI pipeline.
We propose \emph{Bidirectional Teacher (BT)}, including a degradation network $g_{\phi}$ and a recovery network $g_{\psi}$, to store the implicit degradation knowledge.
BT is trained on clean-degraded image pairs, which are often generated by data generation pipelines in CI, \eg~live shooting, or simulation.
In Fig.~\ref{fig:ciada}(a)(b), the pre-trained BT is applied for data transformation, producing augmented data $x_{S_d}$ for the source domain and pseudo-ground-truth $x_{T_{pgt}}$ for the target domain.
Ultimately, we have the source dataset $\mathcal{D}_{S}{=}\{{x_S}^{(i)}, {x_{S_d}}^{(i)}, {y_S}^{(i)}\}_{i=1}^{N_S}$ and target dataset $\mathcal{D}_{T}{=}\{{x_{T_{pgt}}}^{(i)}, {x_T}^{(i)}, {{y_T}_{pl}}^{(i)} \}_{i=1}^{N_T}$ for CIADA training. 
BT enables the auxiliary image recovery task for aberration images in both domains, and the generated data contains the image degradation knowledge from CI data pairs.

\begin{table*}[t]
    \begin{center}
        \caption{Evaluation in mIoU (\%) of classical segmenters on our benchmark. We use segmenters properly trained only on clear Cityscapes from mmsegmentation~\cite{mmseg2020} as KITTI-360 is seldom used for training. The validation images of Cityscapes are resized to $1024{\times}512$. Further implementation details are shown in the Appendix. We highlight the \textbf{best} and \underline{second best} results.}
        \label{tab:sota_seg}
        \renewcommand{\arraystretch}{1.2}
\resizebox{\textwidth}{!}{
\setlength{\tabcolsep}{1mm}{
\begin{tabular}{l|ccccccccccc|ccccccccccc}

\hline
& \multicolumn{11}{c|}{\textbf{Cityscapes}} & \multicolumn{11}{c}{\textbf{KITTI-360}} \\
\cline{2-23}
&Clear&\multicolumn{5}{|c|}{CSL}&\multicolumn{5}{c|}{HRDL}&Clear&\multicolumn{5}{|c|}{CSL}&\multicolumn{5}{c}{HRDL}\\
\cline{2-23}
&-&\multicolumn{1}{|c}{C1}&C2&C3&C4&\multicolumn{1}{c|}{C5}&H1&H2&H3&H4&H5&-&\multicolumn{1}{|c}{C1}&C2&C3&C4&\multicolumn{1}{c|}{C5}&H1&H2&H3&H4&H5 \\
\hline\hline
FCN~\cite{long2015fully} &75.44&74.57&73.23&68.08&37.93&63.20&74.63&73.13&67.40&19.36&52.99&46.44&45.63&42.88&32.80&14.73&31.28&45.64&42.61&27.71&05.02&29.81\\

\hline
PSPNet~\cite{zhao2017pyramid} &75.76&75.08&73.98&69.00&41.41&64.91&75.13&73.97&68.25&21.43&53.35&48.77&48.38&44.42&33.07&13.78&29.20&48.18&44.11&27.73&06.20&25.88\\

\hline
DeepLabV3+~\cite{chen2018encoder} &77.11&76.65&\underline{75.43}&\underline{70.03}&44.18&66.50&76.64&75.32&\underline{69.18}&20.84&54.80&50.38&49.87&47.67&35.50&14.34&31.08&49.65&47.19&29.78&02.22&26.97\\

\hline
SETR~\cite{zheng2021rethinking} &71.47&70.66&69.15&65.57&\textbf{54.05}&64.33&70.98&70.07&66.51&\textbf{49.03}&\underline{63.18}&54.37&\underline{55.62}&\underline{54.71}&\textbf{51.55}&\textbf{36.12}&\textbf{50.20}&\underline{55.78}&\textbf{54.45}&\textbf{51.92}&\textbf{30.20}&\textbf{44.15}\\

\hline
SegFormer~\cite{xie2021segformer} &\underline{78.04}&\underline{77.00}&75.37&69.20&\underline{50.12}&\textbf{67.73}&\underline{77.26}&\underline{76.06}&68.88&\underline{43.67}&\textbf{65.55}&\textbf{59.81}&\textbf{59.03}&\textbf{55.99}&\underline{45.68}&\underline{27.25}&\underline{43.67}&\textbf{57.83}&\underline{53.55}&\underline{45.97}&\underline{17.97}&\underline{42.01}\\

\hline
SegNeXt~\cite{guo2022segnext} &\textbf{79.07}&\textbf{78.62}&\textbf{77.05}&\textbf{70.03}&47.98&\underline{66.99}&\textbf{78.56}&\textbf{77.01}&\textbf{70.12}&37.60&62.93&\underline{56.30}&53.75&52.07&40.55&18.91&36.52&54.06&51.51&37.01&08.38&30.55\\

\hline

\end{tabular}
}
}
    \end{center}
\end{table*}

\PAR{Network architecture.} The network of CIADA contains a shared encoder $E_{\Theta_e}$, a decoder $D_{\Theta_s}$ for segmentation, and an auxiliary decoder $D_{\Theta_r}$ for image recovery, as shown in Fig.~\ref{fig:ciada}(a)(c).
For segmentation against image degradation, we adopt the MiT encoder~\cite{xie2021segformer}, which shows excellent zero-shot robustness on corrupted images, as $E_{\Theta_e}$.
Meanwhile, considering the corruptions on local details of images induced by aberrations, the Hamburger decoder~\cite{guo2022segnext, geng2021attention} is applied for global information extraction in  $D_{\Theta_s}$ and $D_{\Theta_r}$. 
Moreover, except for the mentioned specific architectures, CIADA serves as a plug-and-play distillation framework, which can be applied to any other competitive encoder and decoder.   

\PAR{Correlation-based knowledge distillation.}
Considering the differences between high-level semantic segmentation and low-level image recovery, the single shared encoder architecture hinders the effective interaction of knowledge. 
In CIADA, we propose \emph{Correlation-based Knowledge Distillation} to enhance the distillation process. We introduce the Correlation-based Distillation Loss function (CD Loss) to enforce segmenter $(E_{\Theta_e}, D_{\Theta_s})$ to learn the degradation features extracted in recovery model $(E_{\Theta_e}, D_{\Theta_r})$.
To be specific, we calculate the self-correlation matrix $\mathcal{C}_{s,r}\in\mathbb{R}^{W'H'{\times}H'W'} $ for feature maps ${{F}_{s,r}}\in\mathbb{R}^{C'{\times}H'{\times}W'}$ from decoder to model the associations among all pixels:
\begin{equation}
\label{eq:correlation}
\mathcal{C}_{s,r} = {(\frac{\mathcal{F}_{s,r}}{{\parallel}\mathcal{F}_{s,r}{\parallel}_2})^T}{\cdot}{(\frac{\mathcal{F}_{s,r}}{{\parallel}\mathcal{F}_{s,r}{\parallel}_2})},
\end{equation}
and
\begin{equation}
\label{eq:F}
\mathcal{F}_{s,r} = Reshape(Conv_{1\times 1}(F_{s,r})),
\end{equation}
where $\mathcal{F}_{s,r}\in\mathbb{R}^{C'{\times}H'W'}$ denotes the transformed feature map. The $1{\times}1$ convolution and $L_2$ norm are applied for stable training. Finally, the CD Loss is calculated based on the Charbonnier loss:
\begin{equation}
\label{eq:loss}
\mathcal{L}_{CD}(F_s, F_r) = \sqrt{{\parallel}\mathcal{C}_s-\mathcal{C}_r{\parallel}^2 + \varepsilon^2},
\end{equation}
where $\varepsilon$ is a constant value $10^{-3}$.
The presented CD Loss regularizes features learned from high-level and low-level tasks, distilling the knowledge of how an image degrades with aberrations to the segmenter.

\begin{table*}[t]
\begin{center}
\caption{Comparison with classical unsupervised domain adaptation methods.}
\vskip-2ex
\label{tab:sota_uda}
\renewcommand{\arraystretch}{1.2}
\resizebox{\textwidth}{!}
{
\setlength{\tabcolsep}{1mm}{
\begin{tabular}{ccccccccccccccccccccc}

\hline

\multicolumn{1}{c|}{}&Road&S.walk& Build.&Wall&Fence&Pole&Tr.Light&Sign&Veget.&Terrain&Sky&Person&Rider&Car&Truck&Bus&Train&M.bike&\multicolumn{1}{c|}{Bike}&mIoU\\
\hline\hline
\multicolumn{21}{c}{\textbf{{Cityscapes} $\to$ {KITTI-360-ab (C5) }}}\\
\hline
\multicolumn{1}{c|}{CRST~\cite{zou2019crst}}&69.99&14.54&52.31&21.31&22.50&15.12&00.01&20.23&64.71&42.15&76.22&00.34&00.78&54.80&20.13&01.45&00.00&17.84&\multicolumn{1}{c|}{12.01}&26.25\\
\hline
\multicolumn{1}{c|}{CLAN~\cite{luo2019clan}}&58.54&22.45&67.76&13.43&19.77&16.48&00.87&26.36&76.07&40.95&80.40&18.19&18.17&57.93&14.79&14.00&00.29&10.52&\multicolumn{1}{c|}{04.64}&29.56\\
\hline
\multicolumn{1}{c|}{BDL~\cite{li2019bdl}}&64.38&32.16&68.06&18.13&21.72&13.33&00.09&11.59&75.56&44.77&81.37&06.51&02.70&66.36&23.27&17.10&00.01&24.94&\multicolumn{1}{c|}{05.32}&30.39\\
\hline
\multicolumn{1}{c|}{DACS~\cite{tranheden2021dacs}}&\textbf{84.89}&\textbf{50.14}&74.10&34.37&28.59&19.99&00.10&27.25&80.13&\textbf{48.97}&87.24&10.66&06.08&78.58&16.70&05.65&00.00&\underline{46.03}&\multicolumn{1}{c|}{09.35}&37.31\\
\hline
\multicolumn{1}{c|}{DAFormer~\cite{hoyer2022daformer}}&74.54&38.99&78.44&33.46&\underline{34.98}&\textbf{23.91}&00.66&38.66&\underline{80.54}&46.03&\underline{88.01}&\underline{42.33}&\underline{23.83}&\underline{82.19}&54.09&38.07&\underline{23.63}&42.37&\multicolumn{1}{c|}{\textbf{27.39}}&\underline{45.90}\\
\hline
\multicolumn{1}{c|}{HRDA~\cite{hoyer2022hrda}}&60.10&30.12&\textbf{80.65}&\textbf{41.74}&\textbf{35.63}&19.41&00.43&\textbf{39.17}&78.70&39.67&\textbf{88.96}&27.74&10.40&78.80&\textbf{58.51}&\textbf{40.40}&03.91&43.42&\multicolumn{1}{c|}{23.63}&42.18\\
\hline
\rowcolor{gray!15}\multicolumn{1}{c|}{\textbf{CIADA (Ours)}}&\underline{84.23}&\underline{47.30}&\underline{79.31}&\underline{38.86}&33.07&\underline{22.45}&00.85&\underline{38.70}&\textbf{81.69}&\underline{47.62}&87.34&\textbf{47.60}&\textbf{28.09}&\textbf{83.14}&\underline{56.32}&\underline{39.01}&\textbf{40.46}&\textbf{54.39}&\multicolumn{1}{c|}{\underline{24.15}}&\textbf{49.19}\\
\hline
\multicolumn{21}{c}{\textbf{{Cityscapes} $\to$ {KITTI-360-ab (H5) }}}\\
\hline
\multicolumn{1}{c|}{CRST~\cite{zou2019crst}}&64.34&14.58&48.60&21.18&21.93&17.79&00.00&19.37&62.40&44.35&64.69&0.20&12.96&47.51&10.02&10.42&00.00&12.71&\multicolumn{1}{c|}{16.66}&25.77\\
\hline
\multicolumn{1}{c|}{CLAN~\cite{luo2019clan}}&57.76&18.58&63.59&10.85&19.99&15.59&00.35&23.04&72.61&40.74&73.56&13.27&\textbf{30.03}&53.20&05.59&50.81&00.72&08.33&\multicolumn{1}{c|}{13.22}&30.10\\
\hline
\multicolumn{1}{c|}{BDL~\cite{li2019bdl}}&56.44&26.39&70.29&18.89&20.99&12.14&00.11&06.59&76.39&38.45&81.23&05.91&04.03&64.46&26.78&00.28&00.73&15.69&\multicolumn{1}{c|}{04.64}&27.92\\
\hline
\multicolumn{1}{c|}{DACS~\cite{tranheden2021dacs}}&68.82&35.04&73.51&26.57&28.40&\underline{21.29}&00.03&29.71&\underline{80.09}&\textbf{54.27}&83.96&24.97&17.71&77.53&21.79&56.81&00.91&26.03&\multicolumn{1}{c|}{19.71}&39.32\\
\hline
\multicolumn{1}{c|}{DAFormer~\cite{hoyer2022daformer}}&\underline{69.09}&\underline{35.20}&75.01&\textbf{42.18}&\underline{32.25}&14.73&00.32&\underline{32.82}&78.43&40.74&\underline{86.18}&\textbf{47.30}&\underline{22.69}&\underline{79.43}&32.98&76.72&\underline{08.89}&\textbf{32.98}&\multicolumn{1}{c|}{\textbf{26.71}}&\underline{43.93}\\
\hline
\multicolumn{1}{c|}{HRDA~\cite{hoyer2022hrda}}&57.64&30.51&\underline{75.92}&37.25&\textbf{33.14}&17.26&00.31&\textbf{33.72}&76.64&37.34&84.59&36.67&20.04&69.12&\textbf{39.26}&\textbf{87.62}&06.54&\underline{30.99}&\multicolumn{1}{c|}{18.36}&41.71\\
\hline
\rowcolor{gray!15}\multicolumn{1}{c|}{\textbf{CIADA (Ours)}}&\textbf{77.03}&\textbf{39.52}&\textbf{78.30}&\underline{38.80}&28.26&\textbf{22.95}&00.54&32.42&\textbf{80.35}&\underline{52.83}&\textbf{87.02}&\underline{46.15}&21.55&\textbf{83.84}&\underline{38.31}&\underline{86.69}&\textbf{41.22}&29.25&\multicolumn{1}{c|}{\underline{26.07}}&\textbf{47.95}\\
\hline
\end{tabular}
}
}

\end{center}
\end{table*}

\subsection{Training Pipeline for CIADA}
\label{sec:training}
Based on the ST-UDA baseline, the joint self-training of semantic segmentation and image recovery on images with aberrations is utilized in CIADA:

\PAR{Step 0: Training of BT.}
BT ($g_{\phi}, g_{\psi}$) is first trained for the generation of the degraded source image $x_{S_d}{=}g_{\phi}({x_S})$ and the recovered target image $x_{T_{pgt}}{=}g_{\psi}({x_T})$.
Before CIADA training, the BT is pre-trained, whereas the generation is synchronized with each iteration of the training process. 

\PAR{Step 1: Joint training on source images.} Labeled data $\{x_{S_d}, {y_S}\}$ and paired degraded-clear images $\{x_{S_d}, {x_S}\}$ are applied for source domain training.
We use a combination loss function, composed of categorical Cross-Entropy (CE) loss for semantic segmentation $\mathcal{L}_{CE}$, Charbonnier loss for image recovery $\mathcal{L}_{Char}$, and the proposed CD Loss $\mathcal{L}_{{CD}}$, to supervise the training of the shared encoder architecture:
\begin{equation}
\begin{split}
\label{eq:source}
{\mathcal{L}}_{S} &= \mathcal{L}_{CE}(D_{\Theta_s}(E_{\Theta_e}(x_{S_d})),{y_S}) \\ 
& + w_1\mathcal{L}_{Char}(D_{\Theta_r}(E_{\Theta_e}(x_{S_d})),{x_S}) \\
& + w_2\mathcal{L}_{{CD}}({F_s},{F_r}).
\end{split}
\end{equation}

\PAR{Step2: Generation of pseudo-labels.} We generate pseudo-label ${y_T}_{pl}$ and pseudo weight $q_T$ for the target domain through the teacher segmenter $(E_{\Psi_e},D_{\Psi_s} )$ in accordance with the ST-UDA baseline.  

\PAR{Step 3: Joint training on target images.} With pseudo-label $({y_T}_{pl},q_T)$ and the pseudo-ground-truth image $x_{T_{pgt}}$, we jointly train the network on target data in a similar way: 
\begin{equation}
\begin{split}
\label{eq:t}
{\mathcal{L}}_{T} &= w_3\mathcal{L}_{CE}(D_{\Theta_s}(E_{\Theta_e}(x_{T})), {q_T} {{y_T}_{pl}})\\
& + w_4\mathcal{L}_{Char}(D_{\Theta_r}(E_{\Theta_e}(x_{T})),x_{T_{pgt}}) \\
& + w_5\mathcal{L}_{{CD}}({F_s},{F_r}).
\end{split}
\end{equation}

\PAR{Step 4: Training segmenter on mixed data.}
Class mixing~\cite{tranheden2021dacs} is applied to generate mixed data of both domains: $\{x_M=mix(x_{S_d}, x_T), y_M=mix(y_S,{q_T}{{y_T}_{pl}})\}$. 
The segmenter is trained as:
\begin{equation}
\label{eq:m}
{\mathcal{L}}_{M} = \mathcal{L}_{CE}(D_{\Theta_s}(E_{\Theta_e}(x_{M})), y_M).
\end{equation}

In summary, the entire training objective is:
\begin{equation}
\label{eq:all}
{\mathcal{L}} = \mathcal{L}_{S} + \mathcal{L}_{T} + \mathcal{L}_{M}.
\end{equation}
The $w_1$-$w_5$ in Eq.~\ref{eq:source} and Eq.~\ref{eq:t} are set as $0.05$, $1.00$, $0.01$, $0.05$, and $1.00$ for stable training empirically.
In addition, we follow~\cite{hoyer2022daformer} to update the teacher segmenter for online pseudo-label generation via exponentially moving average~\cite{tarvainen2017mean}.
The training pipeline for CIADA prompts the segmenter to learn how an image degrades with aberrations implicitly from prior knowledge of CI and perform semantic segmentation on unseen images with optical aberrations. 

\section{Experiments}
\label{sec:exp}
We conduct a comprehensive set of experiments to explore SSOA. The implementation details are depicted in Sec.~\ref{sec:setup}. Then, we investigate the robustness of classical segmenters to different distributions of aberrations in Sec.~\ref{sec:influence}. Based on the observations, representative solutions to SSOA are further evaluated in Sec.~\ref{sec:sota} and ablation studies on CIADA are conducted in Sec.~\ref{sec:ablation}.

\subsection{Implementation Details}
For brevity, all of the following settings are consistent for CSL and HRDL, and we will not discuss them separately. All the training processes of our work are implemented on a single NVIDIA GeForce RTX 3090. More details are in the Appendix.  
\label{sec:setup}

\PAR{Dataset for training BT.}
We select the training/test datasets of $1922/447$ ground-truth images from Flickr2K~\cite{timofte2017ntire}, where images are resized to $512{\times}1024$ considering the resolution of images in segmentation datasets.
The simulation pipeline in~\cite{10021856} is applied to generate training data pairs based on the aberration distribution of target MOS images.
In our case, the MOS is VPL groups, whose parameters for PSF generation are known.
To simulate the synthetic-to-real gap, we fine-tune the range of parameters and reconstruct other $20$ distributions. Please refer to the Appendix for more details.  

\PAR{Dataset for optimizing UDA.}
In our benchmark, Cityscapes~\cite{cordts2016cityscapes} ($2975/500$ images for training/validation set with a resolution of $2048{\times}1024$) and KITTI-360~\cite{liao2022kitti} ($3063/768$ images for training/validation set with a resolution of $1408{\times}376$) are used as the source domain dataset, while the corresponding target domains are KITTI-360-ab and Cityscapes-ab, respectively.
The frame rate of KITTI-360 is lowered by $16$ times (detailed in the Appendix), to produce a single-frame dataset with a moderate size and high scene diversity like Cityscapes.
We resize images from Cityscapes to $1024{\times}512$ following a common practice~\cite{hoyer2022daformer}. 

\PAR{Training.}
We use an efficient image restoration baseline~\cite{chen2022simple} for networks in BT. 
The degradation- and the recovery network are trained separately.
By default, we train BT according to~\cite{chen2022simple}. 
We follow~\cite{hoyer2022daformer} to train ST-UDA, with a batch size $2$ on random crops of $376{\times}376$ for $40K$ iterations.

\subsection{Influences of Optical Aberrations}
\label{sec:influence}

The previous work~\cite{kamann2020benchmarking} reveals that most segmenters are robust to aberration-based PSF blurs of conventional optical systems. 
However, the influences of much more severe aberrations of MOS have not been investigated. 
Based on the benchmark, we evaluate classical segmenters in TABLE~\ref{tab:sota_seg} under different behaviors and levels of aberrations. 
All the segmenters are robust to slight aberrations (C1/H1 and C2/H2) consistent with~\cite{kamann2020benchmarking} but suffer considerable destructive effects when the aberration is severe, especially for level C4/H4 (${-}17.42{\sim}{-}48.18$ in mIoU). 
Compared to CNN-based models~\cite{chen2018encoder,guo2022segnext,long2015fully,zhao2017pyramid}, transformer-based segmenters~\cite{xie2021segformer,zheng2021rethinking} are much more robust to aberrations which have higher mIoU under comprehensive sets of C5/H5 and severe aberrations sets of C4/H4, despite that the CNN-based SegNeXt~\cite{guo2022segnext} shows superior performance on clear images. In summary, for MOS whose aberrations are often severe, the influence of optical aberrations should be considered prudently for its applications in semantic segmentation. Further, the transformer architecture could be essential for SSOA with MOS due to its inherent robustness to image corruption. 

\subsection{Comparison with Possible Solutions}
\label{sec:sota}
\PAR{UDA methods.} 
Without access to labeled aberration images, UDA is a preferred solution to SSOA.
TABLE~\ref{tab:sota_uda} shows per-class results of previous UDA methods (including designs that improve the robustness, such as data augmentation, robust network architecture, and adversarial learning) and our proposed CIADA under our UDA Setting I. 
All models are retrained on $376{\times}376$ random crops with a batch size of $2$ for $40K$ iterations.
The transformer architecture proves to be crucial to SSOA for the considerably much higher performance of DAFormer, HRDA, and CIADA.
Moreover, compared to the baseline method DAFormer~\cite{hoyer2022daformer}, CIADA improves mIoU from $45.90$ to $49.19$ under CSL and $43.93$ to $47.95$ under HRDL due to the distillation of prior knowledge of image degradation from CI pipeline. 

\begin{table}[!t]
\begin{center}
\caption{Evaluation in mIoU (\%) of possible solutions to semantic segmentation under different aberrations distributions.}
\label{tab:sota_all}

\renewcommand{\arraystretch}{1.2}
\resizebox{0.5\textwidth}{!}{
\setlength{\tabcolsep}{1mm}{    
\begin{tabular}{l|ccccc|ccccc}

\hline

&\multicolumn{5}{c|}{CSL}&\multicolumn{5}{c}{HRDL}\\
\cline{2-11}
&C1&C2&C3&C4&C5&H1&H2&H3&H4&H5\\
\cline{2-11}
&\multicolumn{5}{c|}{\textbf{{Cs.} $\to$ {KI.-ab} (C5)}}&\multicolumn{5}{c}{\textbf{{Cs} $\to$ {KI.-ab} (H5)}} \\

\hline\hline
Oracle &{59.98}&{56.80}&{53.32}&{43.93}&{53.57}&{59.25}&{58.34}&{50.57}&{40.06}&{55.62}\\

\hline
Src-Only &52.15&49.84&38.77&19.11&37.60&52.76&49.96&35.47&11.49&36.91\\

\hline
CI\&Seg &54.00&51.64&43.82&23.50&41.21&53.30&51.91&42.40&15.78&37.37\\

\hline
ST-UDA&54.86&53.41&46.41&30.17&45.90&53.50&52.33&46.01&19.19&43.93\\

\hline
\textbf{CIADA (Ours)}&\cellcolor{gray!15}{56.28}&\cellcolor{gray!15}{55.17}&\cellcolor{gray!15}{49.20}&\cellcolor{gray!15}{32.29}&\cellcolor{gray!15}{49.19}&\cellcolor{gray!15}{56.64}&\cellcolor{gray!15}{55.34}&\cellcolor{gray!15}{47.45}&\cellcolor{gray!15}{21.24}&\cellcolor{gray!15}{47.95}\\

\hline
&\multicolumn{5}{c|}{\textbf{{KI.} $\to$ {Cs.-ab} (C5)}}&\multicolumn{5}{c}{\textbf{{KI.} $\to$ {Cs.-ab} (H5)}}\\

\hline\hline
Oracle &{72.95}&{72.72}&{69.78}&{58.31}&{68.03}&{72.16}&{71.88}&{68.89}&{55.27}&{66.20}\\

\hline
Src-Only &50.77&48.02&41.45&28.60&41.82&50.87&48.21&40.04&17.19&36.20\\

\hline
CI\&Seg &51.88&51.09&48.56&36.82&47.20&52.39&51.83&45.25&31.30&44.54\\

\hline
ST-UDA&58.71&58.08&55.34&45.61&54.49&60.21&59.23&55.47&35.38&51.89\\

\hline
\textbf{CIADA (Ours)}&\cellcolor{gray!15}{60.24}&\cellcolor{gray!15}{59.97}&\cellcolor{gray!15}{57.79}&\cellcolor{gray!15}{49.11}&\cellcolor{gray!15}{56.77}&\cellcolor{gray!15}{60.59}&\cellcolor{gray!15}{60.20}&\cellcolor{gray!15}{57.56}&\cellcolor{gray!15}{45.58}&\cellcolor{gray!15}{55.37}\\
\hline
\end{tabular}
}
}

\end{center}
\end{table}

\begin{table}[!t]
\begin{center}
\caption{Evaluation in mIoU (\%) of CI\&Seg solution with more recovery networks.}
\label{tab:ciseg}
\resizebox{0.5\textwidth}{!}{
\setlength{\tabcolsep}{1mm}{    
\begin{tabular}{l|ccccc|ccccc}

\hline

&\multicolumn{5}{c|}{CSL}&\multicolumn{5}{c}{HRDL}\\
\cline{2-11}
&C1&C2&C3&C4&C5&H1&H2&H3&H4&H5\\
\cline{2-11}
&\multicolumn{5}{c|}{\textbf{{Cs.} $\to$ {KI.-ab} (C5)}}&\multicolumn{5}{c}{\textbf{{Cs} $\to$ {KI.-ab} (H5)}} \\

\hline\hline
\textbf{CIADA (Ours)}&\cellcolor{gray!15}{56.28}&\cellcolor{gray!15}{55.17}&\cellcolor{gray!15}{49.20}&\cellcolor{gray!15}{32.29}&\cellcolor{gray!15}{49.19}&\cellcolor{gray!15}{56.64}&\cellcolor{gray!15}{55.34}&\cellcolor{gray!15}{47.45}&\cellcolor{gray!15}{21.24}&\cellcolor{gray!15}{47.95}\\

\hline
Src-Only &52.15&49.84&38.77&19.11&37.60&52.76&49.96&35.47&11.49&36.91\\

\hline
CI\&Seg (NAFNet~\cite{chen2022simple}) &54.00&51.64&43.82&23.50&41.21&53.30&51.91&42.40&15.78&37.37\\

\hline
CI\&Seg (Restormer~\cite{zamir2022restormer}) &53.32&51.16&44.90&24.77&41.63&53.47&51.84&43.92&18.85&41.83\\

\hline
CI\&Seg (KPN~\cite{mildenhall2018burst})&51.88&50.01&42.62&24.76&39.96&53.36&49.65&39.60&13.44&36.56\\

\hline
&\multicolumn{5}{c|}{\textbf{{KI.} $\to$ {Cs.-ab} (C5)}}&\multicolumn{5}{c}{\textbf{{KI.} $\to$ {Cs.-ab} (H5)}}\\

\hline\hline
\textbf{CIADA (Ours)}&\cellcolor{gray!15}{60.24}&\cellcolor{gray!15}{59.97}&\cellcolor{gray!15}{57.79}&\cellcolor{gray!15}{49.11}&\cellcolor{gray!15}{56.77}&\cellcolor{gray!15}{60.59}&\cellcolor{gray!15}{60.20}&\cellcolor{gray!15}{57.56}&\cellcolor{gray!15}{45.58}&\cellcolor{gray!15}{55.37}\\

\hline
Src-Only &50.77&48.02&41.45&28.60&41.82&50.87&48.21&40.04&17.19&36.20\\

\hline
CI\&Seg (NAFNet~\cite{chen2022simple}) &51.88&51.09&48.56&36.82&47.20&52.39&51.83&45.25&31.30&44.54\\

\hline
CI\&Seg (Restormer~\cite{zamir2022restormer}) &52.03&51.63&49.11&35.56&47.19&52.44&52.03&45.20&24.85&41.91\\

\hline
CI\&Seg (KPN~\cite{mildenhall2018burst}) &52.08&50.97&45.29&31.49&44.99&52.84&51.83&42.70&24.43&40.63\\

\hline

\hline
\end{tabular}
}
}
\end{center}
\end{table}

\begin{figure*}[!t]
  \centering
  \includegraphics[width=1.0\linewidth]{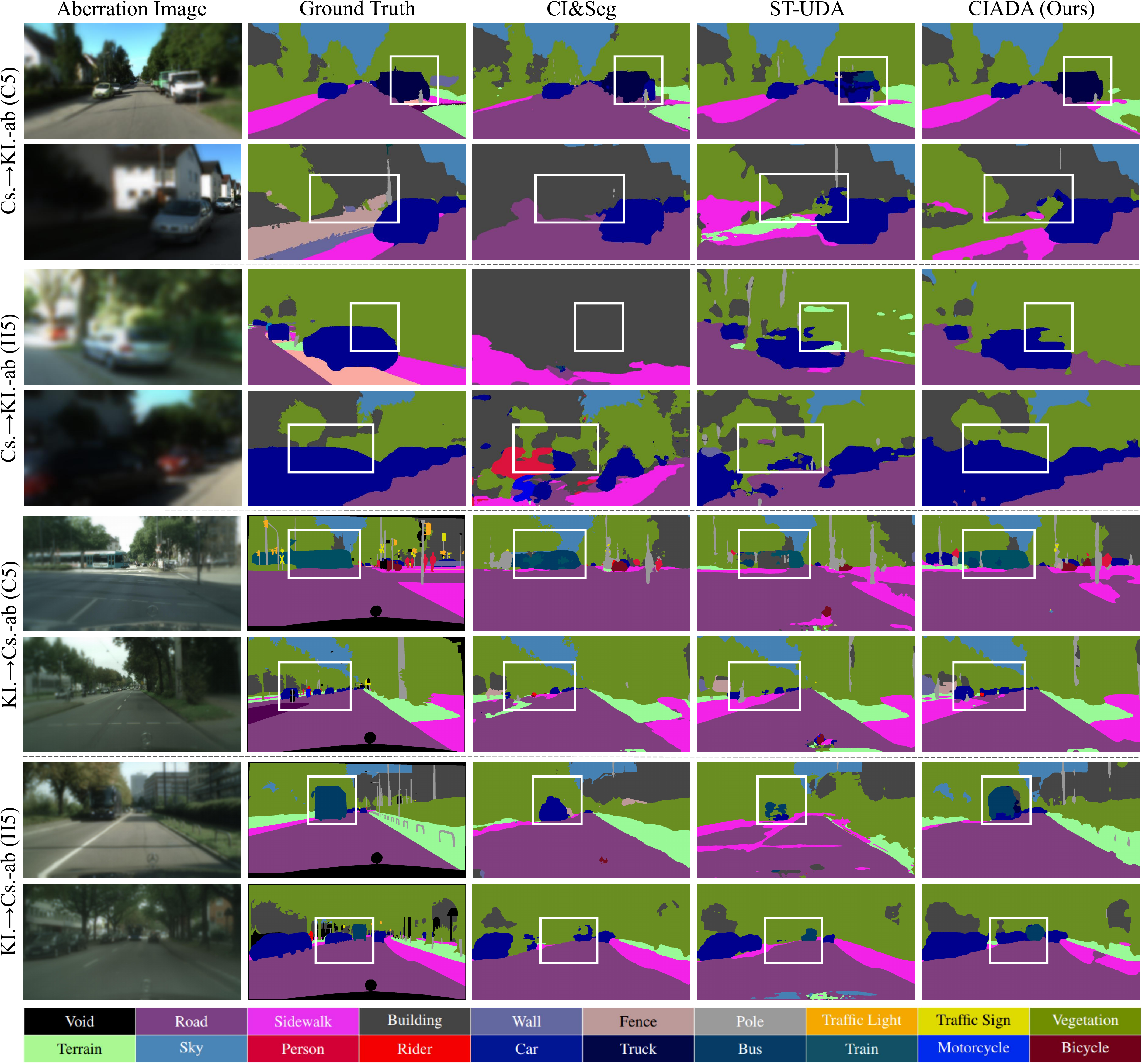}
  \caption{Qualitative performance of possible solutions to SSOA. We compare our solution, CIADA, with other competitive pipelines, \ie~CI\&Seg (image recovery before segmentation) and ST-UDA (our baseline, DAFormer). CIADA reveals superiority in all settings, whose visual results are relatively clean, coherent, and accurate (highlighted with white boxes).}
  \label{fig:quality}
\end{figure*}

\PAR{Possible solutions to SSOA.} We evaluate possible solutions to SSOA based on our benchmark, \ie~Oracle (trained with labeled target data), Src-Only (trained only with labeled source data), CI\&Seg (recovery before segmentation), ST-UDA, and our proposed CIADA. 
SegFormer~\cite{xie2021segformer} is adopted for the Oracle and Src-Only training, while DAFormer~\cite{hoyer2022daformer} is applied for ST-UDA, for their robustness and efficiency. The recovery network in BT is applied before the Src-Only model in CI\&Seg for a fair comparison.
TABLE~\ref{tab:sota_all} shows the mIoU of each solution under all behaviors and levels of aberrations. 
The corresponding qualitative results are also shown in Fig.~\ref{fig:quality}, where some special classes of ground truth are not considered during the training process as a common practice~\cite{hoyer2022daformer,hoyer2022hrda}. 

For Src-Only methods, the performances drop dramatically as the aberration level becomes more severe.
Although with extra computational overhead, the cascade of the recovery model and the Src-Only model only brings limited improvements, which is unable to produce convincing scene understanding results as is shown in Fig.~\ref{fig:quality}. 
ST-UDA improves the segmentation results significantly, where most of the classes in the scene can be understood roughly.
However, there exist terrible noises in its visual segmentation results and the segmentation of each instance is incoherent, especially at the junction of instances.
CIADA proves to be a superior solution to SSOA in terms of achieving the best mIoU score in all settings, whose visual results are also relatively clean, coherent, and accurate.
The relative performance and the results under rPC metric in Fig.~\ref{fig:solution} also illustrate that CIADA can effectively mitigate the gap between the clear domain and the aberration domain.
The usage of knowledge in CI helps the segmenter learn how the image degrades, which is essential to SSOA.

\begin{table*}[!t]
	\renewcommand\arraystretch{1.2}
	\setlength{\fboxrule}{0pt}
	\begin{center}
		
\begin{minipage}[ht]{0.3\textwidth}
	\begin{center}
	    \caption{Ablations on CI knowledge.}
	    \vspace{-3pt}
		\setlength{\tabcolsep}{2mm}
		\resizebox{\textwidth}{!}{
            \begin{tabular}{c|cc|c}
            
            \bottomrule

            \textbf{Solution}&\textbf{Degradation}&\textbf{Recovery}&\textbf{mIOU}\\
            \hline\hline
            \multirow{4}{*}

            {{CIADA}}&-&-&45.07\\
            \cline{2-4}
            &$\surd$&-&47.25\\
            \cline{2-4}
            &-&$\surd$&47.24\\
            \cline{2-4}
           &\cellcolor{gray!15}$\surd$&\cellcolor{gray!15}$\surd$&\cellcolor{gray!15}\textbf{49.19}\\
            \hline
            \multirow{1}{*}{Src-only}&\multicolumn{2}{c|}{-}&37.60\\
            \cline{2-4}

            \toprule
            \end{tabular}
		}
	    \label{tab_ab1}
	\end{center}
\end{minipage}
\hspace{2pt}
\begin{minipage}[ht]{0.32\textwidth}
	\begin{center}
		\caption{Ablations on decoders.}
            \label{tab_ab2}

		\vspace{-4pt}
		\setlength{\tabcolsep}{2mm}
		\resizebox{\textwidth}{!}{
            \begin{tabular}{c|ccc}
            
            \bottomrule

            \textbf{Decoder}&\textbf{Src-Only}&\textbf{ST-UDA}&\textbf{CIADA}\\
            \hline\hline
            UperNet~\cite{xiao2018unified}&36.60&\textbf{46.74}&47.83\\
            \cline{1-4}
            ASPP~\cite{chen2018encoder}&38.61&46.48&47.99\\
            \cline{1-4}
            SegFormer~\cite{xie2021segformer}&37.60&45.11&46.37\\
            \cline{1-4}
            DAFormer~\cite{hoyer2022daformer}&\textbf{38.85}&45.90&46.90\\
            \cline{1-4}
            Hamburger$^*$~\cite{guo2022segnext}&35.09&44.69&48.23\\
            \cline{1-4}
            \rowcolor{gray!15}Hamburger~\cite{guo2022segnext}&36.40&45.07&\textbf{48.55}\\
            \toprule
            \end{tabular}
		}
		\label{tab:resolution}
	\end{center}
\end{minipage}
\hspace{2pt}
\begin{minipage}[ht]{0.20\textwidth}
	\begin{center}
	    \caption{Ablations on CD Loss.}
	    \vspace{-4pt}
		\setlength{\tabcolsep}{1.0mm}
		\resizebox{\textwidth}{!}{
            \begin{tabular}{c|c}
            
            \bottomrule

            \textbf{Position of CD Loss}&\textbf{mIOU}\\
            \hline\hline
            w/o&48.20\\
            \hline
            \rowcolor{gray!15}burger-in&\textbf{49.19}\\
            \hline
            burger-out&49.01\\
            \hline
            align-out&48.55\\
            \hline
            all&48.71\\
            \toprule
            \end{tabular}
		}
	    \label{tab_ab3}
	\end{center}
\end{minipage}
\hspace{5pt}
	\end{center}
\end{table*}

\PAR{CI$\&$Seg under more recovery models.} We replace the NAFNet in CI\&Seg solution with other image recovery networks for comprehensive evaluations of the commonly acknowledged pipeline for solving the problems of MOS in scene understanding. Restormer~\cite{zamir2022restormer} is selected as a transformer-based image restoration network due to its powerful ability to handle various kinds of image degradation. KPN~\cite{mildenhall2018burst} is applied as a representation of the network designing for processing the spatially-variant degradation, which is often compared in aberration correction~\cite{chen2021optical,10021856}. All the recovery networks are trained on the dataset of BT with the same training details of NAFNet. In accordance with TABLE~\ref{tab:sota_all}, the segmentation models in TABLE~\ref{tab:ciseg} are trained on our four UDA settings and tested on validation sets of different levels of aberrations of CSL and HRDL respectively.    

For image degradation caused by aberrations of MOS, especially severe aberrations, recovering an optimal image for human observation is an ill-posed problem that is hard to solve. Moreover, an optimal recovered image for human observation is not necessarily optimal for downstream tasks. Consequently, as TABLE~\ref{tab:ciseg} shows, applying a CI pipeline before the segmenter brings little improvement to the Src-Only model, where different recovery networks only have limited influence on the results. However, as a knowledge-distillation-based solution, CIADA, which transfers the optimal image degradation information to the target segmenter, can achieve much better performance without the additional computational overhead of the recovery model.

\subsection{Ablation Study}
\label{sec:ablation}
We conduct ablation studies to investigate why the proposed CIADA solution achieves superior SSOA results. The CD Loss is placed before the task head in TABLE~\ref{tab_ab2} considering different decoder architectures. In all cases, the experiments are implemented under {Cityscapes}${\to}${KITTI-360-ab (C5)} and the model is evaluated on the validation set of KITTI-360-ab (C5) in mIoU ($\%$).

\PAR{Ablations on CI knowledge.}
The CI knowledge of applied MOS is stored in the bidirectional teacher.
TABLE~\ref{tab_ab1} shows the results under different usages of BT.
Based on a simple baseline network, the single degradation teacher provides a mIoU increase of $2.18$ due to the data augmentation of aberrations, while the single recovery teacher improves the UDA performance by $2.17$ in mIoU through implicit knowledge distillation of the auxiliary recovery in the target domain. 
The combination of two teachers achieves the highest gain of $4.12$, which enables auxiliary recovery learning in both domains. 

\PAR{Ablations on decoders.}
TABLE~\ref{tab_ab2} shows that Hamburger decoder~\cite{guo2022segnext} achieves the best CIADA result of $48.55$ in mIoU. 
Compared to the simple SegFormer decoder~\cite{xie2021segformer} or other decoders focusing on global context (\ie~UperNet~\cite{xiao2018unified}, ASPP~\cite{chen2018encoder}, and DAFormer~\cite{hoyer2022daformer}), the matrix-decomposition-based Hamburger can learn better global context in both segmentation and recovery~\cite{geng2021attention}, which is essential for CIADA with the auxiliary recovery task.
In addition, the Hamburger aggregating all encoder features (Hamburger$^*$) brings a mIoU drop of $0.32$, showing that the shallow layer features hurt the performance of CIADA. 

\PAR{Ablations on the CD Loss.}
The CD Loss can be applied to any feature in the decoder, where we define burger-in, burger-out, and align-out for the Hamburger decoder in Fig.~\ref{fig:ciada}(c). 
As shown in TABLE~\ref{tab_ab3}, the CD Loss brings improvements of $0.35{\sim}0.99$ due to the knowledge distillation between two tasks.
The optimum position for the CD Loss is burger-in, which is far from the task head.
It reveals that the knowledge is best distilled at the early stage of the decoder due to the differences between the two tasks in CIADA.

\begin{figure*}[t]
  \centering
  \includegraphics[width=1.0\linewidth]{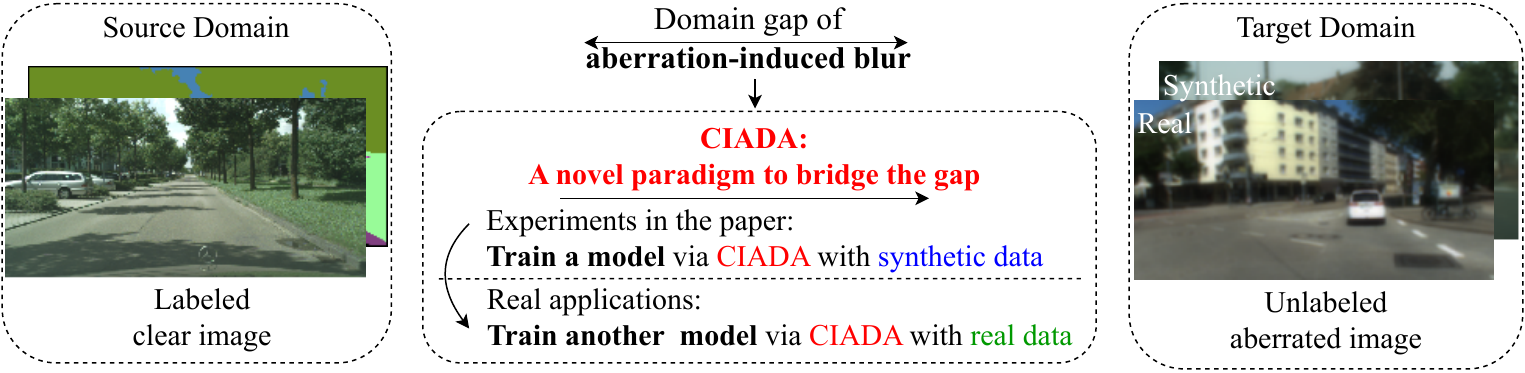}
  \caption{Illustration of how our framework adapts to real data. The UDA setting simulates the real situation in that we can only capture unlabeled aberrated data with the target MOS. The designed experiments on synthetic benchmarks are to illustrate the ability of a UDA framework in SSOA tasks with unlabeled aberration data, which is a reflection of how well it works on real MOS.}
  \label{fig:reb}
\end{figure*}
\section{Conclusion and Discussion}
\label{sec:clu}
\subsection{Conclusion}
In this paper, the conventional computational imaging for human perception is extended to computational imaging for machine perception on mobile agents, boosting the applications of MOS in downstream tasks.
We look into Semantic Segmentation under Optical Aberrations (SSOA) via an unsupervised domain adaptation perspective, which deals with the synthetic-to-real gap in conventional supervised settings. 
We put forward Virtual Prototype Lens (VPL) groups to generate datasets under different aberration distributions for benchmarking SSOA, where UDA tasks are set up without access to the target labels. Additionally, the CIADA framework is proposed for robust SSOA with implicit prior knowledge of CI via multi-task learning. Experimental results reveal that our CIADA is a superior solution, which improves the performance by $3.29/4.02$ and $2.28/3.48$ in mIoU over the baseline ST-UDA framework in two UDA settings.
Moreover, in terms of the framework design, CIADA is a successful attempt to combine advanced CI techniques with SSOA without additional computation overhead during inference.

\subsection{Discussion and Future Work}

Our work appears to be a landmark for investigating the applications of MOS in semantic scene understanding. It is the first work to address semantic segmentation under severe optical aberrations of MOS to the
best of our knowledge. However, some limitations and future work require to be further addressed for constructing a more solid training and evaluation system.

\PAR{Further performance improvement space.} 
There remains considerable improvement space in the performance of solutions to SSOA, both in mIoU and qualitative results. CIADA and ST-UDA can make the segmenter successfully understand the scene roughly under aberrations, but the results of details, especially the joint of some adjacent instances, are unsatisfactory, which makes it disadvantageous for applications of MOS in semantic scene understanding. To deal with the tough task, we believe that more stages and iterations of training with a larger batch size are worth trying to reach robust performance. The auxiliary CI pipeline can also be applied to other UDA frameworks for exploring better results, which will benefit from the blooming development of UDA methods.

\PAR{The synthetic-to-real gap.} 
Our work carefully addresses the synthetic-to-real gap problem via setting up an unsupervised domain adaptive SSOA setting. 
For an intuitive setting, when training a segmenter with simulated aberration images and corresponding labels, the differences between simulated images and real captured MOS images induce the synthetic-to-real gap.
In this setting, when applying the trained model on real MOS, the gap will influence the prospective performance, leading to the unconvincing synthetic benchmark for SSOA.
However, in our UDA setting, the training process has no access to the labels of the target data.
In this case, the situations of real and synthetic data are equivalent, as both processes can be interpreted as domain adaptation from labeled clear images to unlabeled aberration images as shown
in Fig.~\ref{fig:reb}.
When applying CIADA in real scenes, some unlabeled images with applied MOS will be collected for training another model.
Thus, the success of CIADA on our benchmark is a reflection of how well it works on real MOS.
A dataset captured by real MOS with diverse aberrations is expected to further benefit the research on SSOA, yet, we point out that our benchmark is a pioneering and convincing effort towards a better understanding of SSOA in real applications, as there are no existing datasets for aberrated images with dense labels. We aim to capture real aberration images by different MOS samples with fine annotations for a more convincing benchmark in our future work.

\PAR{Combinations with neural and differentiable lenses.}
Recent advances in neural and differentiable lenses~\cite{cote2023differentiable,xian2023neural} make it possible to train and optimize optical systems and downstream algorithms together, which can also be understood essentially as high-level vision tasks on aberrated images.
However, these works are based on the differentiable imaging simulation engine~\cite{wang2022differentiable}, which does not consider the errors in lens manufacture and assembly. 
In other words, the potential synthetic-to-real domain gap will be detrimental to the compatibility between the designed lens and downstream models, resulting in performance degradation.
Consequently, the neural and differentiable lenses community will benefit from our work, where the setup UDA framework can be adopted to mitigate the gap and fine-tune the models for promising real-world results.
Our work can also benefit from the neural and differentiable lenses. For example, by using the engine to automatically design MOS in batch, we can generate more realistic MOS samples to improve the diversity of VPLs, for constructing benchmarks with smoother gradation of aberration levels and more diverse aberration distributions. In the future, how to organically combine CIADA with neural and differentiable lenses will be an important and interesting research point.


{\small
\bibliographystyle{IEEEtran}
\bibliography{egbib}

\begin{thebibliography}{10}
\providecommand{\url}[1]{#1}
\csname url@samestyle\endcsname
\providecommand{\newblock}{\relax}
\providecommand{\bibinfo}[2]{#2}
\providecommand{\BIBentrySTDinterwordspacing}{\spaceskip=0pt\relax}
\providecommand{\BIBentryALTinterwordstretchfactor}{4}
\providecommand{\BIBentryALTinterwordspacing}{\spaceskip=\fontdimen2\font plus
\BIBentryALTinterwordstretchfactor\fontdimen3\font minus \fontdimen4\font\relax}
\providecommand{\BIBforeignlanguage}[2]{{%
\expandafter\ifx\csname l@#1\endcsname\relax
\typeout{** WARNING: IEEEtran.bst: No hyphenation pattern has been}%
\typeout{** loaded for the language `#1'. Using the pattern for}%
\typeout{** the default language instead.}%
\else
\language=\csname l@#1\endcsname
\fi
#2}}
\providecommand{\BIBdecl}{\relax}
\BIBdecl

\bibitem{cordts2016cityscapes}
M.~Cordts \emph{et~al.}, ``The cityscapes dataset for semantic urban scene understanding,'' in \emph{Proc. CVPR}, 2016, pp. 3213--3223.

\bibitem{zhang2022trans4trans}
J.~Zhang, K.~Yang, A.~Constantinescu, K.~Peng, K.~M{\"u}ller, and R.~Stiefelhagen, ``{Trans4Trans:} {Efficient} transformer for transparent object and semantic scene segmentation in real-world navigation assistance,'' \emph{{IEEE} Trans. Intell. Transp. Syst.}, vol.~23, no.~10, pp. 19\,173--19\,186, 2022.

\bibitem{le2022bayesian}
H.~T. Le, S.~L. Phung, and A.~Bouzerdoum, ``Bayesian gabor network with uncertainty estimation for pedestrian lane detection in assistive navigation,'' \emph{{IEEE} Trans. Circuits Syst. Video Technol.}, vol.~32, no.~8, pp. 5331--5345, 2022.

\bibitem{yang2018unifying}
K.~Yang, L.~M. Bergasa, E.~Romera, R.~Cheng, T.~Chen, and K.~Wang, ``Unifying terrain awareness through real-time semantic segmentation,'' in \emph{Proc. IV}, 2018, pp. 1033--1038.

\bibitem{liu2023open}
R.~Liu \emph{et~al.}, ``Open scene understanding: Grounded situation recognition meets segment anything for helping people with visual impairments,'' in \emph{Proc. ICCVW}, 2023, pp. 1849--1859.

\bibitem{cui2020development}
W.~Cui, C.~Chang, and L.~Gao, ``Development of an ultra-compact optical combiner for augmented reality using geometric phase lenses,'' \emph{Opt. Lett.}, vol.~45, no.~10, pp. 2808--2811, 2020.

\bibitem{sun2021aerial}
L.~Sun \emph{et~al.}, ``{Aerial-PASS:} {Panoramic} annular scene segmentation in drone videos,'' in \emph{Proc. ECMR}, 2021, pp. 1--6.

\bibitem{wang2022high}
J.~Wang \emph{et~al.}, ``High-performance panoramic annular lens design for real-time semantic segmentation on aerial imagery,'' \emph{Opt. Eng.}, vol.~61, no.~3, pp. 035\,101--035\,101, 2022.

\bibitem{zhang2023large}
Y.~Zhang \emph{et~al.}, ``Large depth-of-field ultra-compact microscope by progressive optimization and deep learning,'' \emph{Nat. Commun.}, vol.~14, no.~1, p. 4118, 2023.

\bibitem{uddin2016search}
Z.~Uddin and M.~Islam, ``Search and rescue system for alive human detection by semi-autonomous mobile rescue robot,'' in \emph{Proc. ICISET}, 2016, pp. 1--5.

\bibitem{deng2020semantic}
W.~Deng \emph{et~al.}, ``{Semantic RGB-D SLAM} for rescue robot navigation,'' \emph{IEEE Access}, vol.~8, pp. 221\,320--221\,329, 2020.

\bibitem{hua2022ultra}
X.~Hua \emph{et~al.}, ``Ultra-compact snapshot spectral light-field imaging,'' \emph{Nat. Commun.}, vol.~13, no.~1, p. 2732, 2022.

\bibitem{liu2022computational}
K.~Liu \emph{et~al.}, ``Computational imaging for simultaneous image restoration and super-resolution image reconstruction of single-lens diffractive optical system,'' \emph{Appl. Sci.}, vol.~12, no.~9, p. 4753, 2022.

\bibitem{xie2021segformer}
E.~Xie, W.~Wang, Z.~Yu, A.~Anandkumar, J.~M. Alvarez, and P.~Luo, ``{SegFormer:} {Simple} and efficient design for semantic segmentation with transformers,'' in \emph{Proc. NeurIPS}, vol.~34, 2021, pp. 12\,077--12\,090.

\bibitem{chen2022simple}
L.~Chen, X.~Chu, X.~Zhang, and J.~Sun, ``Simple baselines for image restoration,'' in \emph{Proc. ECCV}, vol. 13667, 2022, pp. 17--33.

\bibitem{michaelis2019benchmarking}
C.~Michaelis \emph{et~al.}, ``Benchmarking robustness in object detection: Autonomous driving when winter is coming,'' \emph{arXiv preprint arXiv:1907.07484}, 2019.

\bibitem{hummer2023vltseg}
C.~H{\"u}mmer, M.~Schwonberg, L.~Zhong, H.~Cao, A.~Knoll, and H.~Gottschalk, ``{VLTSeg:} {Simple} transfer of {CLIP-based} vision-language representations for domain generalized semantic segmentation,'' \emph{arXiv preprint arXiv:2312.02021}, 2023.

\bibitem{dun2020learned}
X.~Dun, H.~Ikoma, G.~Wetzstein, Z.~Wang, X.~Cheng, and Y.~Peng, ``Learned rotationally symmetric diffractive achromat for full-spectrum computational imaging,'' \emph{Optica}, vol.~7, no.~8, pp. 913--922, 2020.

\bibitem{Peng2019LearnedLF}
Y.~Peng, Q.~Sun, X.~Dun, G.~Wetzstein, W.~Heidrich, and F.~Heide, ``Learned large field-of-view imaging with thin-plate optics,'' \emph{{ACM} Trans. Graph.}, vol.~38, no.~6, pp. 1--14, 2019.

\bibitem{diamond2021dirty}
S.~Diamond, V.~Sitzmann, F.~Julca-Aguilar, S.~Boyd, G.~Wetzstein, and F.~Heide, ``Dirty pixels: Towards end-to-end image processing and perception,'' \emph{{ACM} Trans. Graph.}, vol.~40, no.~3, pp. 1--15, 2021.

\bibitem{diamond2017dirty}
S.~Diamond, V.~Sitzmann, S.~Boyd, G.~Wetzstein, and F.~Heide, ``Dirty pixels: Optimizing image classification architectures for raw sensor data,'' \emph{arXiv preprint arXiv:1701.06487}, 2017.

\bibitem{sakaridis2018semantic}
C.~Sakaridis, D.~Dai, and L.~Van~Gool, ``Semantic foggy scene understanding with synthetic data,'' \emph{Int. J. Comput. Vis.}, vol. 126, no.~9, pp. 973--992, 2018.

\bibitem{pei2019effects}
Y.~Pei, Y.~Huang, Q.~Zou, X.~Zhang, and S.~Wang, ``Effects of image degradation and degradation removal to {CNN-based} image classification,'' \emph{{IEEE} Trans. Pattern Anal. Mach. Intell.}, vol.~43, no.~4, pp. 1239--1253, 2021.

\bibitem{vidalmata2020bridging}
R.~G. VidalMata \emph{et~al.}, ``Bridging the gap between computational photography and visual recognition,'' \emph{{IEEE} Trans. Pattern Anal. Mach. Intell.}, vol.~43, no.~12, pp. 4272--4290, 2021.

\bibitem{liao2022kitti}
Y.~Liao, J.~Xie, and A.~Geiger, ``{KITTI-360:} {A} novel dataset and benchmarks for urban scene understanding in {2D} and {3D},'' \emph{{IEEE} Trans. Pattern Anal. Mach. Intell.}, vol.~45, no.~3, pp. 3292--3310, 2023.

\bibitem{hoyer2022daformer}
L.~Hoyer, D.~Dai, and L.~Van~Gool, ``{DAFormer:} {Improving} network architectures and training strategies for domain-adaptive semantic segmentation,'' in \emph{Proc. CVPR}, 2022, pp. 9914--9925.

\bibitem{luo2019clan}
Y.~Luo, L.~Zheng, T.~Guan, J.~Yu, and Y.~Yang, ``Taking a closer look at domain shift: Category-level adversaries for semantics consistent domain adaptation,'' in \emph{Proc. CVPR}, 2019, pp. 2507--2516.

\bibitem{zou2019crst}
Y.~Zou, Z.~Yu, X.~Liu, B.~V. K.~V. Kumar, and J.~Wang, ``Confidence regularized self-training,'' in \emph{Proc. ICCV}, 2019, pp. 5981--5990.

\bibitem{mahajan1994zernike}
V.~N. Mahajan, ``Zernike circle polynomials and optical aberrations of systems with circular pupils,'' \emph{Appl. Opt.}, vol.~33, no.~34, pp. 8121--8124, 1994.

\bibitem{barbastathis2019use}
G.~Barbastathis, A.~Ozcan, and G.~Situ, ``On the use of deep learning for computational imaging,'' \emph{Optica}, vol.~6, no.~8, pp. 921--943, 2019.

\bibitem{schuler2011non}
C.~J. Schuler, M.~Hirsch, S.~Harmeling, and B.~Sch{\"o}lkopf, ``Non-stationary correction of optical aberrations,'' in \emph{Proc. ICCV}, 2011, pp. 659--666.

\bibitem{wu2020non}
X.~Wu, H.~Yang, B.~Liu, and X.~Liu, ``Non-uniform deblurring for simple lenses imaging system,'' in \emph{Proc. AEMCSE}, 2020, pp. 274--278.

\bibitem{zamir2022restormer}
S.~W. Zamir, A.~Arora, S.~Khan, M.~Hayat, F.~S. Khan, and M.-H. Yang, ``Restormer: Efficient transformer for high-resolution image restoration,'' in \emph{Proc. CVPR}, 2022, pp. 5718--5729.

\bibitem{chen2021optical}
S.~Chen, H.~Feng, D.~Pan, Z.~Xu, Q.~Li, and Y.~Chen, ``Optical aberrations correction in postprocessing using imaging simulation,'' \emph{{ACM} Trans. Graph.}, vol.~40, no.~5, pp. 1--15, 2021.

\bibitem{10021856}
Q.~Jiang, H.~Shi, L.~Sun, S.~Gao, K.~Yang, and K.~Wang, ``Annular computational imaging: Capture clear panoramic images through simple lens,'' \emph{IEEE Trans. Comput. Imaging}, vol.~8, pp. 1250--1264, 2022.

\bibitem{li2021universal}
X.~Li, J.~Suo, W.~Zhang, X.~Yuan, and Q.~Dai, ``Universal and flexible optical aberration correction using deep-prior based deconvolution,'' in \emph{Proc. ICCV}, 2021, pp. 2593--2601.

\bibitem{peng2016diffractive}
Y.~Peng, Q.~Fu, F.~Heide, and W.~Heidrich, ``The diffractive achromat full spectrum computational imaging with diffractive optics,'' \emph{{ACM} Trans. Graph.}, vol.~35, no.~4, pp. 1--11, 2016.

\bibitem{chen2018encoder}
L.-C. Chen, Y.~Zhu, G.~Papandreou, F.~Schroff, and H.~Adam, ``Encoder-decoder with atrous separable convolution for semantic image segmentation,'' in \emph{Proc. ECCV}, vol. 11211, 2018, pp. 833--851.

\bibitem{guo2022segnext}
M.-H. Guo, C.-Z. Lu, Q.~Hou, Z.~Liu, M.-M. Cheng, and S.-M. Hu, ``{SegNeXt:} {Rethinking} convolutional attention design for semantic segmentation,'' in \emph{Proc. NeurIPS}, vol.~35, 2022, pp. 1140--1156.

\bibitem{long2015fully}
J.~Long, E.~Shelhamer, and T.~Darrell, ``Fully convolutional networks for semantic segmentation,'' in \emph{Proc. CVPR}, 2015, pp. 3431--3440.

\bibitem{guo2019degraded}
D.~Guo, Y.~Pei, K.~Zheng, H.~Yu, Y.~Lu, and S.~Wang, ``Degraded image semantic segmentation with dense-gram networks,'' \emph{{IEEE} Trans. Image Process.}, vol.~29, pp. 782--795, 2020.

\bibitem{kamann2020increasing_painting}
C.~Kamann and C.~Rother, ``Increasing the robustness of semantic segmentation models with painting-by-numbers,'' in \emph{Proc. ECCV}, vol. 12355, 2020, pp. 369--387.

\bibitem{romera2019bridging}
E.~Romera, L.~M. Bergasa, K.~Yang, J.~M. Alvarez, and R.~Barea, ``Bridging the day and night domain gap for semantic segmentation,'' in \emph{Proc. IV}, 2019, pp. 1312--1318.

\bibitem{liu2023improving_nighttime}
W.~Liu, W.~Li, J.~Zhu, M.~Cui, X.~Xie, and L.~Zhang, ``Improving nighttime driving-scene segmentation via dual image-adaptive learnable filters,'' \emph{{IEEE} Trans. Circuits Syst. Video Technol.}, 2023.

\bibitem{sakaridis2021acdc}
C.~Sakaridis, D.~Dai, and L.~Van~Gool, ``{ACDC:} {The} adverse conditions dataset with correspondences for semantic driving scene understanding,'' in \emph{Proc. ICCV}, 2021, pp. 10\,745--10\,755.

\bibitem{zendel2018wilddash}
O.~Zendel, K.~Honauer, M.~Murschitz, D.~Steininger, and G.~F. Dominguez, ``{WildDash} - {Creating} hazard-aware benchmarks,'' in \emph{ECCV}, vol. 11210, 2018, pp. 407--421.

\bibitem{kamann2020benchmarking}
C.~Kamann and C.~Rother, ``Benchmarking the robustness of semantic segmentation models,'' in \emph{Proc. CVPR}, 2020, pp. 8825--8835.

\bibitem{zhou2022fan}
D.~Zhou \emph{et~al.}, ``Understanding the robustness in vision transformers,'' in \emph{Proc. ICML}, vol. 162, 2022, pp. 27\,378--27\,394.

\bibitem{muller2023classification}
P.~M{\"u}ller, A.~Braun, and M.~Keuper, ``Classification robustness to common optical aberrations,'' in \emph{Proc. ICCV}, 2023, pp. 3634--3645.

\bibitem{lian2019constructing_self_motivated}
Q.~Lian, F.~Lv, L.~Duan, and B.~Gong, ``Constructing self-motivated pyramid curriculums for cross-domain semantic segmentation: A non-adversarial approach,'' in \emph{Proc. ICCV}, 2019, pp. 6757--6766.

\bibitem{zhang2017curriculum_domain_adaptation}
Y.~Zhang, P.~David, and B.~Gong, ``Curriculum domain adaptation for semantic segmentation of urban scenes,'' in \emph{Proc. ICCV}, 2017, pp. 2039--2049.

\bibitem{9785619}
Y.~Zhao, Z.~Zhong, Z.~Luo, G.~H. Lee, and N.~Sebe, ``Source-free open compound domain adaptation in semantic segmentation,'' \emph{{IEEE} Trans. Circuits Syst. Video Technol.}, vol.~32, no.~10, pp. 7019--7032, 2022.

\bibitem{chang2019all_about_structure}
W.-L. Chang, H.-P. Wang, W.-H. Peng, and W.-C. Chiu, ``All about structure: Adapting structural information across domains for boosting semantic segmentation,'' in \emph{Proc. CVPR}, 2019, pp. 1900--1909.

\bibitem{tsai2018adaptsegnet}
Y.-H. Tsai, W.-C. Hung, S.~Schulter, K.~Sohn, M.-H. Yang, and M.~Chandraker, ``Learning to adapt structured output space for semantic segmentation,'' in \emph{Proc. CVPR}, 2018, pp. 7472--7481.

\bibitem{9889741}
C.~Cao \emph{et~al.}, ``Adversarial dual-student with differentiable spatial warping for semi-supervised semantic segmentation,'' \emph{{IEEE} Trans. Circuits Syst. Video Technol.}, vol.~33, no.~2, pp. 793--803, 2023.

\bibitem{goodfellow2020gan}
I.~Goodfellow \emph{et~al.}, ``Generative adversarial networks,'' \emph{Commun. {ACM}}, vol.~63, no.~11, pp. 139--144, 2020.

\bibitem{huang2020contextual_relation_consistent}
J.~Huang, S.~Lu, D.~Guan, and X.~Zhang, ``Contextual-relation consistent domain adaptation for semantic segmentation,'' in \emph{Proc. ECCV}, vol. 12360, 2020, pp. 705--722.

\bibitem{li2020content_consistent}
G.~Li, G.~Kang, W.~Liu, Y.~Wei, and Y.~Yang, ``Content-consistent matching for domain adaptive semantic segmentation,'' in \emph{Proc. ECCV}, vol. 12359, 2020, pp. 440--456.

\bibitem{wang2020differential_treatment}
Z.~Wang \emph{et~al.}, ``Differential treatment for stuff and things: A simple unsupervised domain adaptation method for semantic segmentation,'' in \emph{Proc. CVPR}, 2020, pp. 12\,632--12\,641.

\bibitem{zheng2021rectifying_pseudo_label}
Z.~Zheng and Y.~Yang, ``Rectifying pseudo label learning via uncertainty estimation for domain adaptive semantic segmentation,'' \emph{Int. J. Comput. Vis.}, vol. 129, no.~4, pp. 1106--1120, 2021.

\bibitem{yang2020fda}
Y.~Yang and S.~Soatto, ``{FDA:} {Fourier} domain adaptation for semantic segmentation,'' in \emph{Proc. CVPR}, 2020, pp. 4084--4094.

\bibitem{kang2020pixel_cycle_association}
G.~Kang, Y.~Wei, Y.~Yang, Y.~Zhuang, and A.~Hauptmann, ``Pixel-level cycle association: A new perspective for domain adaptive semantic segmentation,'' in \emph{Proc. NeurIPS}, vol.~33, 2020, pp. 3569--3580.

\bibitem{vu2019advent}
T.-H. Vu, H.~Jain, M.~Bucher, M.~Cord, and P.~P{\'e}rez, ``{ADVENT:} {Adversarial} entropy minimization for domain adaptation in semantic segmentation,'' in \emph{Proc. CVPR}, 2019, pp. 2517--2526.

\bibitem{zhang2021proda}
P.~Zhang, B.~Zhang, T.~Zhang, D.~Chen, Y.~Wang, and F.~Wen, ``Prototypical pseudo label denoising and target structure learning for domain adaptive semantic segmentation,'' in \emph{Proc. CVPR}, 2021, pp. 12\,414--12\,424.

\bibitem{xie2022sepico}
B.~Xie, S.~Li, M.~Li, C.~H. Liu, G.~Huang, and G.~Wang, ``{SePiCo:} {Semantic-guided} pixel contrast for domain adaptive semantic segmentation,'' \emph{{IEEE} Trans. Pattern Anal. Mach. Intell.}, vol.~45, no.~7, pp. 9004--9021, 2023.

\bibitem{tranheden2021dacs}
W.~Tranheden, V.~Olsson, J.~Pinto, and L.~Svensson, ``{DACS:} {Domain} adaptation via cross-domain mixed sampling,'' in \emph{Proc. WACV}, 2021, pp. 1378--1388.

\bibitem{9889681}
Q.~Zhou \emph{et~al.}, ``Context-aware mixup for domain adaptive semantic segmentation,'' \emph{{IEEE} Trans. Circuits Syst. Video Technol.}, vol.~33, no.~2, pp. 804--817, 2023.

\bibitem{hoyer2022hrda}
L.~Hoyer, D.~Dai, and L.~Van~Gool, ``{HRDA:} {Context-aware} high-resolution domain-adaptive semantic segmentation,'' in \emph{Proc. ECCV}, vol. 13690, 2022, pp. 372--391.

\bibitem{huggins2007introduction}
E.~Huggins, ``Introduction to fourier optics,'' \emph{Phys. Teach.}, vol.~45, no.~6, pp. 364--368, 2007.

\bibitem{araslanov2021self_supervised_augmentation_consistency}
N.~Araslanov and S.~Roth, ``Self-supervised augmentation consistency for adapting semantic segmentation,'' in \emph{Proc. CVPR}, 2021, pp. 15\,384--15\,394.

\bibitem{huo2022domain_agnostic_prior}
X.~Huo, L.~Xie, H.~Hu, W.~Zhou, H.~Li, and Q.~Tian, ``Domain-agnostic prior for transfer semantic segmentation,'' in \emph{Proc. CVPR}, 2022, pp. 7065--7075.

\bibitem{lai2022decouplenet}
X.~Lai \emph{et~al.}, ``{DecoupleNet:} {Decoupled} network for domain adaptive semantic segmentation,'' in \emph{Proc. ECCV}, vol. 13693, 2022, pp. 369--387.

\bibitem{li2022class_balanced_pixel_level_self_labeling}
R.~Li, S.~Li, C.~He, Y.~Zhang, X.~Jia, and L.~Zhang, ``Class-balanced pixel-level self-labeling for domain adaptive semantic segmentation,'' in \emph{Proc. CVPR}, 2022, pp. 11\,583--11\,593.

\bibitem{wang2020dual}
L.~Wang, D.~Li, Y.~Zhu, L.~Tian, and Y.~Shan, ``Dual super-resolution learning for semantic segmentation,'' in \emph{Proc. CVPR}, 2020, pp. 3773--3782.

\bibitem{mmseg2020}
M.~Contributors, ``{MMSegmentation}: {OpenMMLab} semantic segmentation toolbox and benchmark,'' \url{https://github.com/open-mmlab/mmsegmentation}, 2020.

\bibitem{zhao2017pyramid}
H.~Zhao, J.~Shi, X.~Qi, X.~Wang, and J.~Jia, ``Pyramid scene parsing network,'' in \emph{Proc. CVPR}, 2017, pp. 6230--6239.

\bibitem{zheng2021rethinking}
S.~Zheng \emph{et~al.}, ``Rethinking semantic segmentation from a sequence-to-sequence perspective with transformers,'' in \emph{Proc. CVPR}, 2021, pp. 6881--6890.

\bibitem{geng2021attention}
Z.~Geng, M.-H. Guo, H.~Chen, X.~Li, K.~Wei, and Z.~Lin, ``Is attention better than matrix decomposition?'' in \emph{Proc. ICLR}, 2021.

\bibitem{li2019bdl}
Y.~Li, L.~Yuan, and N.~Vasconcelos, ``Bidirectional learning for domain adaptation of semantic segmentation,'' in \emph{Proc. CVPR}, 2019, pp. 6936--6945.

\bibitem{tarvainen2017mean}
A.~Tarvainen and H.~Valpola, ``Mean teachers are better role models: Weight-averaged consistency targets improve semi-supervised deep learning results,'' in \emph{Proc. NeurIPS}, vol.~30, 2017, pp. 1195--1204.

\bibitem{timofte2017ntire}
R.~Timofte \emph{et~al.}, ``{NTIRE} 2017 challenge on single image super-resolution: Methods and results,'' in \emph{Proc. CVPRW}, 2017, pp. 1110--1121.

\bibitem{mildenhall2018burst}
B.~Mildenhall, J.~T. Barron, J.~Chen, D.~Sharlet, R.~Ng, and R.~Carroll, ``Burst denoising with kernel prediction networks,'' in \emph{Proc. CVPR}, 2018, pp. 2502--2510.

\bibitem{xiao2018unified}
T.~Xiao, Y.~Liu, B.~Zhou, Y.~Jiang, and J.~Sun, ``Unified perceptual parsing for scene understanding,'' in \emph{Proc. ECCV}, vol. 11209, 2018, pp. 432--448.

\bibitem{cote2023differentiable}
G.~C{\^o}t{\'e}, F.~Mannan, S.~Thibault, J.-F. Lalonde, and F.~Heide, ``The differentiable lens: Compound lens search over glass surfaces and materials for object detection,'' in \emph{Proc. CVPR}, 2023, pp. 20\,803--20\,812.

\bibitem{xian2023neural}
W.~Xian, A.~Bo{\v{z}}i{\v{c}}, N.~Snavely, and C.~Lassner, ``Neural lens modeling,'' in \emph{Proc. CVPR}, 2023, pp. 8435--8445.

\bibitem{wang2022differentiable}
C.~Wang, N.~Chen, and W.~Heidrich, ``{dO:} {A} differentiable engine for deep lens design of computational imaging systems,'' \emph{IEEE Trans. Computational Imaging}, vol.~8, pp. 905--916, 2022.

\bibitem{basicsr}
X.~Wang, L.~Xie, K.~Yu, K.~C. Chan, C.~C. Loy, and C.~Dong, ``{BasicSR}: Open source image and video restoration toolbox,'' \url{https://github.com/XPixelGroup/BasicSR}, 2022.

\bibitem{loshchilov2017decoupled}
I.~Loshchilov and F.~Hutter, ``Decoupled weight decay regularization,'' in \emph{ICLR}, 2019.

\bibitem{loshchilov2016sgdr}
------, ``{SGDR:} {Stochastic} gradient descent with warm restarts,'' in \emph{ICLR}, 2017.

\bibitem{beyer2023flexivit}
L.~Beyer \emph{et~al.}, ``{FlexiViT:} {One} model for all patch sizes,'' in \emph{Proc. CVPR}, 2023, pp. 14\,496--14\,506.

\bibitem{tian2023resformer}
R.~Tian, Z.~Wu, Q.~Dai, H.~Hu, Y.~Qiao, and Y.-G. Jiang, ``{ResFormer:} {Scaling} {ViTs} with multi-resolution training,'' in \emph{Proc. CVPR}, 2023, pp. 22\,721--22\,731.

\end{thebibliography}
}
%

\appendices
\counterwithin{figure}{section}
\counterwithin{equation}{section}
\counterwithin{table}{section}
\section{Construction of Virtual Prototype Lens}

We supplement the details of the construction of our Virtual Prototype Lens (VPL) in this section.
In Sec.~\ref{sec:pipeline}, we illustrate the wave-based theory for imaging simulation of MOS.
The statistics of MOS samples are detailed in Sec.~\ref{sec:statistics}, based on which we generate our VPL groups in Sec.~\ref{sec:settings}. 

\subsection{Wave-based Imaging Simulation}
\label{sec:pipeline}
Considering the complex and distinct structures of different MOS, we abstract the specific optical system into a black box. By calculating the corresponding Point Spread Function (PSF) of each image patch under different Fields of View (FoVs), we can simulate the imaging result based on patch-wise convolution.

Without loss of generality, we take an imaging system with the circular pupil as an example, and firstly only discuss PSF calculation under a certain fixed FoV $\theta_0$ and wavelength $\lambda_0$. Assuming that a point light source is on the object plane, the corresponding intensity distribution on the image plane after propagation is the PSF. 

Based on Fourier optics, the pupil function on the pupil plane is modulated by aberrations. In this case, the effect of aberrations is to deviate the wavefront at the pupil from the ideal sphere. The wave aberration $W(x,y)$ is applied to represent the optical path difference between the two surfaces, so the phase function $\Phi(x,y)$ on the pupil plane is expressed as:
\begin{equation}
\label{eq:phase}
\Phi(x,y) = k_0W(x,y),
\end{equation}
where $k_0=\frac{2\pi}{\lambda_0}$ denotes the wave number. 

Zernike polynomial is a mathematical description of optical wavefronts propagating through circular pupils~\cite{mahajan1994zernike}. $\Phi$ can therefore be described by Zernike circle polynomials $\mathcal{W}(\rho,\theta)$ in polar coordinates as:
\begin{equation}
\label{eq:zernike_sup}
 \Phi(\rho,\theta) = \mathcal{W}(\rho,\theta) = \sum_{n,m} {C^m_n}{Z^m_n}(\rho,\theta),
\end{equation}
where $C$ denotes Zernike coefficients and $Z$ refers to polynomials. The combination of different $m$ and $n$ represents different orders. The above polar expression of $(\rho,\theta)$ can be rewritten in rectangular coordinates of $(x,y)$ by coordinate transformation ($x=\rho\cos\theta, y=\rho\sin\theta$).

With the phase function, the pupil function $\mathcal{P}(x,y)$ can be written by:
\begin{equation}
\label{eq:pupil}
\mathcal{P}(x,y) = P(x,y)e^{\mathrm{i}\mathcal{W}(x,y)},
\end{equation}
where $P(x,y)$ is circ function in our case. 

In such an imaging system, the light field propagating from the back surface to the image plane satisfies Fresnel diffraction theory and the amplitude can be calculated by scalar diffraction integral~\cite{huggins2007introduction}. Before integral, we distinguish the coordinates of the image plane and the pupil plane: let the coordinate of the image plane be $(x,y)$ and that of the pupil plane be $(x',y')$. Therefore, the amplitude of the image plane $E(x,y)$ can be expressed as:
\begin{equation}
\label{eq:integral}
E(x,y) = \frac{E_0}{{\lambda_0}d}\iint\mathcal{P}(x',y')e^{-\mathrm{i}\frac{2\pi}{{\lambda_0}d}(x'x + y'y)}dx'dy',
\end{equation}
where $E_0$ is a constant amplitude related to illumination, $d$ denotes the distance from the pupil plane to the image plane and $\lambda_0$ is the selected wavelength mentioned above. 

The above discussion is carried out at a certain FoV $\theta_0$ and wavelength $\lambda_0$. In fact, the calculated $E$ on the image plane is a function of FoV $\theta$ and wavelength $\lambda$. More generally, we write Equ.~\eqref{eq:integral} as:
\begin{equation}
\label{eq:integrallambda}
E(x,y,\theta,\lambda) = \frac{E_0}{{\lambda}d}\iint{\mathcal{P}(x',y',\theta,\lambda)e^{-\mathrm{i}\frac{2\pi}{{\lambda}d}(x'x + y'y)}dx'dy'}.
\end{equation}

Finally, the focal intensity on the image plane is:
\begin{equation}
\label{eq:intensity}
I(x,y,\theta,\lambda) = {|E(x,y,\theta,\lambda)|^2}.
\end{equation}
Under our hypothetical point light source condition, the above intensity is the PSF distribution $K(x,y,\theta,\lambda)$ required. In this way, we can get the degraded image patch $I_{ab}(x,y,\theta)$ at FoV $\theta$ induced by aberrations of MOS by: 
\begin{equation}
\label{eq:conv2}
I_{ab}(x,y,\theta) = \int{I_{clear}(x,y,\theta)\otimes K(x,y,\theta,\lambda)d\lambda},
\end{equation}
where $I_{clear}(x,y,\theta)$ is the clear image patch and $\otimes$ denotes convolution.  

\begin{figure*}[!t]
  \centering
  \includegraphics[width=1.0\linewidth]{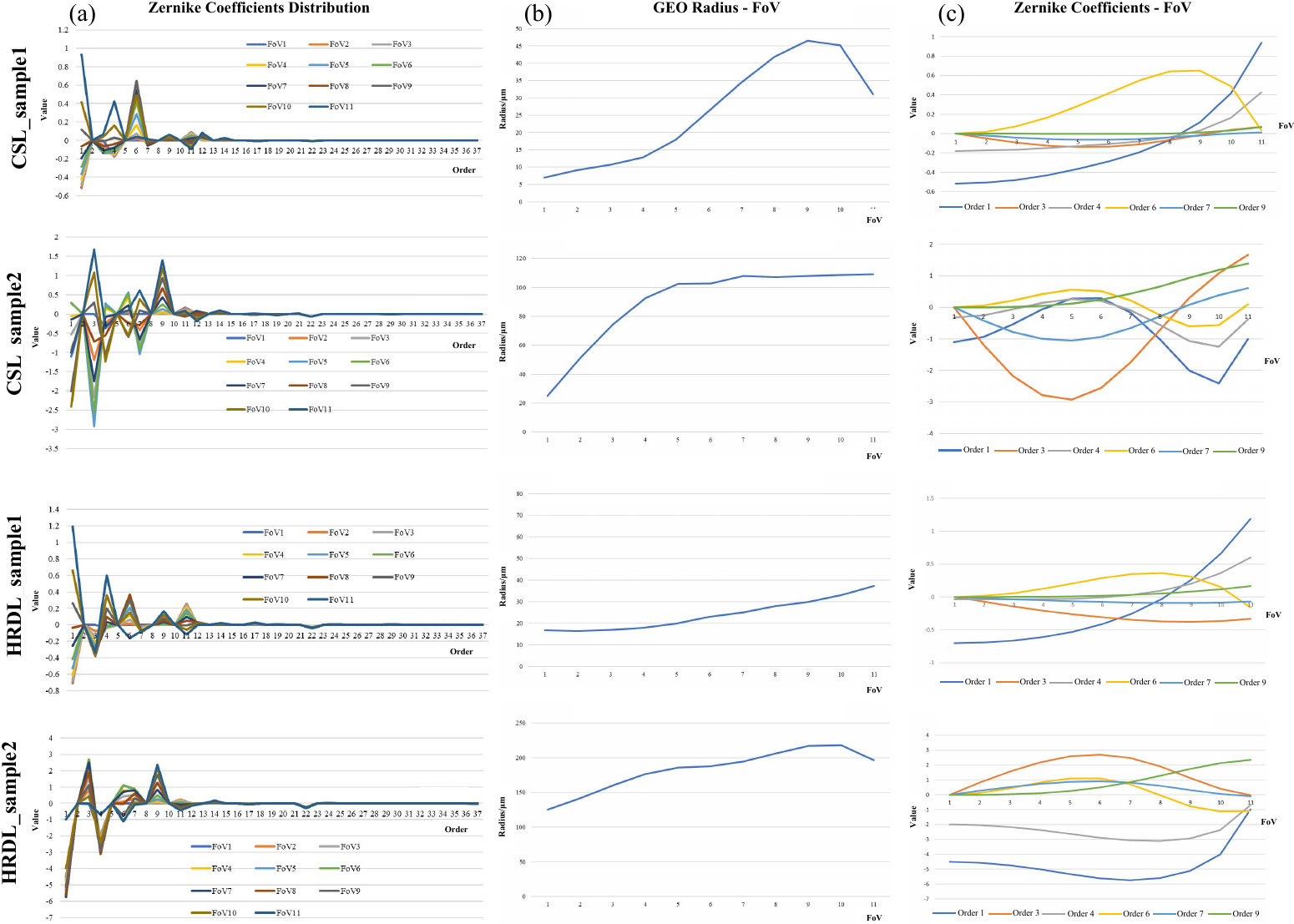}
  \caption{Statistics analysis of MOS samples. Considering the different FoVs of different MOS samples, we regularize the FoVs into $11$ normalized FoVs for all samples. (a) Distributions of Zernike coefficients under $11$ normalized FoVs. (b) Geometrical (GEO) radius distribution over FoVs. (c) Distributions of high impacts Zernike orders over normalized FoVs. The high impacts Zernike orders of MOS samples are: piston (Order1), $y$ tilt (Order3), defocus (Order4), $x$ primary astigmatism (Order6), $y$ primary coma (Order7), and $y$ trefoil (Order9), which are concluded from (a).}
  \label{fig:statistics}
\end{figure*}

\subsection{Statistics Analysis of MOS Samples}
\label{sec:statistics}
To generate virtual MOS samples for establishing the benchmark, we analyze the distribution of Zernike coefficients over FoVs and the radius of spot diagram of different MOS samples, containing samples of Common Simple Lens (CSL) and Hybrid Refractive Diffractive Lens (HRDL).
Fig.~\ref{fig:statistics} shows the statistics results of two CSL samples and two HRDL samples, which can be used to determine the ranges and curve trends of random Zernike coefficients and kernel sizes of PSFs.
Specifically, we conclude the Zernike orders of high values in Fig~\ref{fig:statistics}(a) and analyze the curve trends over FoVs of them in (c), establishing the random range of each high impact order and the type of corresponding curve fitting.
Moreover, the GEO radius-FoV distributions in Fig.~\ref{fig:statistics}(b) are used for setting the size of PSF kernels of different levels and behaviors of aberrations.
It reveals that the GEO radius distribution of HRDL is more uniform than CSL, which represents the spatial-uniform aberration behavior. 

\subsection{Settings of VPL groups}
\label{sec:settings}
For a comprehensive understanding of how optical aberrations affect semantic segmentation, we establish different levels of aberrations distributions: C1-C4 for CSL and H1-H4 for HRDL based on the above statistics.
Additional levels C5 and H5 are hybrid sets of C1-C4 and H1-H4 respectively for an overall evaluation. 

\begin{figure}[h]
  \centering
  \includegraphics[width=1.0\linewidth]{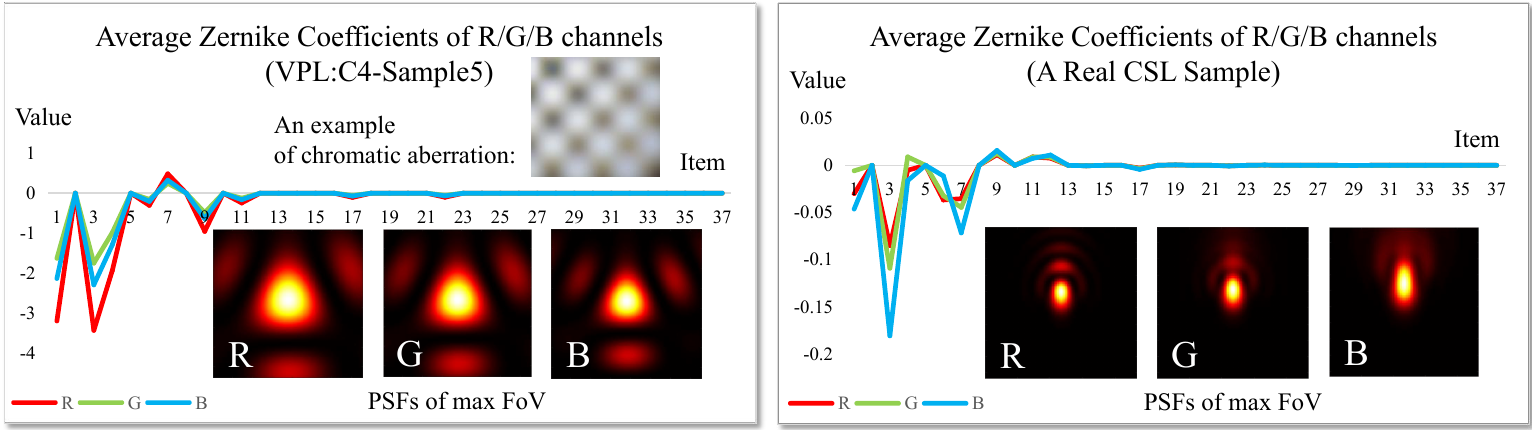}
  \caption{Illustration of chromatic aberrations considered in our VPL groups. Considering the common correction process in MOS design, the chromatic aberrations are set to be small in our benchmark. }
  \label{fig:chroma}
\end{figure}

TABLE~\ref{tab:vpl} shows the random ranges of parameters for constructing VPL groups. $(x,y)$ for Radius means the minimum GEO radius $x$ in the center FoV and maximum one $y$ in the edge FoV. The overall range $(-x,y)$ denotes that the max value of negative coefficients is $-x$ while $y$ is for positive ones.  We generate random Zernike coefficients over $128$ normalized FoVs of each order according to its curve trend. Specifically, take one order for example, we first randomly select a possible curve trend which is concluded in statistics analysis. Then, a random number $r$ is generated among ranges $(x^-,y^-)/(x^+,y^+)$ in the last six columns of TABLE~\ref{tab:vpl}, where the former range is for negative values and the latter one is for positive values. In this way, the peak of the curve is calculated by the product of $r$ and overall range, based on which we fit the coefficients-FoV curves for all $37$ orders and generate a Zernike coefficients matrix of $37{\times}128$. 
Finally, the matrix that represents the aberrations distribution of one virtual MOS sample, \ie~the VPL sample, is used to calculate PSFs for imaging simulation. We point out that the chromatic aberration is considered in the VPL groups via tuning the generated coefficients by $25\%\sim50\%$ for R/G/B channels, which reflect the
result of a combination of different wavelengths.
The average Zernike coefficients and PSFs under $3$ channels of a real CSL sample and a VPL sample are shown in Fig.~\ref{fig:chroma}.
As chromatic aberration would be first corrected to some extent in optical design, the deviates between $3$ channels are set to be small.

\begin{figure*}[!t]
  \centering
  \includegraphics[width=1.0\linewidth]{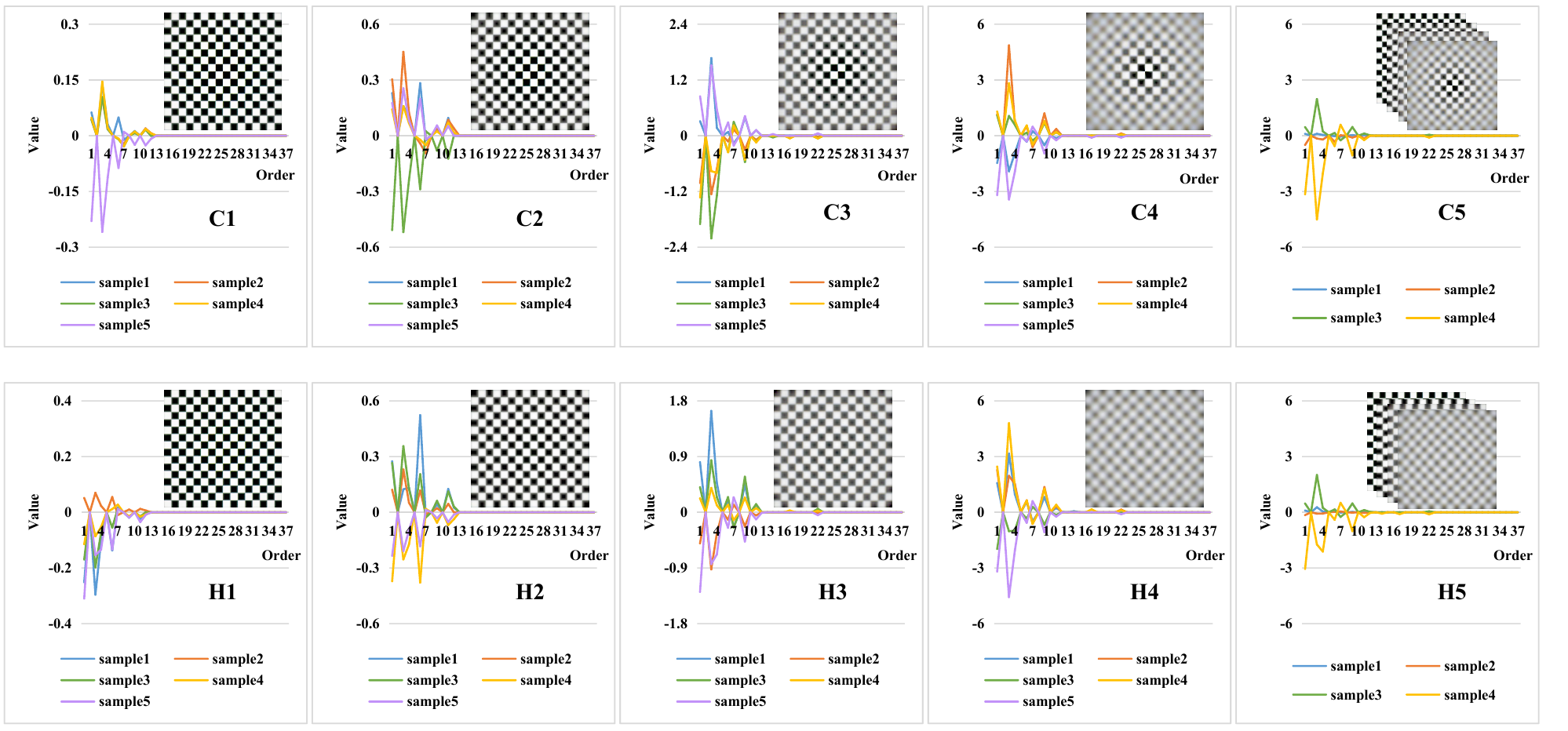}
  \caption{Average Zernike coefficients distributions over FoVs and RGB channels of different samples for each level. We also show the imaging results of the checkerboard of each level. In C5 and H5, we select $4$ samples of different levels for example. Our VPL groups contain diverse aberrations distributions which are significant for benchmarking SSOA.}
  \label{fig:vplsample}
\end{figure*}

Additionally, for a visualized understanding of our VPL samples and aberration levels, we show the imaging results of the checkerboard of each level in Fig.~\ref{fig:vplsample}. The average Zernike coefficients over FoVs and RGB channels of each level are also exhibited (including samples in level 5).
Our VPL with the diverse aberrations distributions is a landmark engine for benchmarking SSOA. 

\begin{table*}[t]
    \begin{center}
        \caption{The ranges for random generation of GEO Radius (/${\mu}m$) and Zernike coefficients. All the random ranges are concluded from chosen MOS samples which ensures that the generated random distributions conform to the aberrations behaviors of most MOS as much as possible. }
        \label{tab:vpl}
        \resizebox{1\textwidth}{!}{
\setlength{\tabcolsep}{1mm}{    
\begin{tabular}{l|cccccccc}

\hline

&\textbf{Radius}&\textbf{Overall}&\multicolumn{6}{c}{\textbf{Specific Orders}}\\
&-&-&1&3&4&6&7&9\\

\hline\hline
C1&(5, 35)&(-0.4, 0.3)&(0.45, 1)/(0.8, 1)&(0.45, 1)/(0.45, 1)&(0.2, 0.5)/(0.3, 0.5)&(0.1, 0.5)/(0.2, 0.7)&(0.1, 0.35)/(0, 0.1)&(0.1, 0.15)/(0.1, 0.15)\\

\hline
C2&(5, 75)&(-0.7, 0.95)&(0.45, 1)/(0.8, 1)&(0.45, 1)/(0.45, 1)&(0.2, 0.5)/(0.3, 0.5)&(0.1, 0.5)/(0.2, 0.7)&(0.1, 0.35)/(0, 0.1)&(0.1, 0.15)/(0.1, 0.15)\\

\hline
C3&(5, 150)&(-3, 1.5)&(0.6, 1)/(0.8, 1)&(0.45, 1)/(0.45, 1)&(0.2, 0.5)/(0.3, 0.5)&(0.1, 0.25)/(0.1, 0.5)&(0.1, 0.2)/(0, 0.2)&(0.45, 0.6)/(0.45, 0.6)\\

\hline
C4&(5, 300)&(-6, 5)&(0.8, 1)/(0.8, 1)&(0.9, 1)/(0.9, 1)&(0.25, 0.7)/(0.4, 0.7)&(0.1, 0.15)/(0.1, 0.3)&(0.1, 0.15)/(0.2, 0.4)&(0.45, 0.8)/(0.45, 0.8)\\

\hline
H1&(15, 25)&(-0.3, 0.3)&(0.3, 0.8)/(0.75, 0.9)&(0.3, 0.8)/(0.6, 0.8)&(0.3, 0.4)/(0.35, 0.55)&(0.1, 0.3)/(0.4, 0.7)&(0.1, 0.2)/(0, 0.15)&(0.1, 0.15)/(0.1, 0.15)\\

\hline
H2&(15, 25)&(-0.6, 0.7)&(0.3, 0.8)/(0.75, 0.9)&(0.3, 0.8)/(0.6, 0.8)&(0.3, 0.4)/(0.35, 0.55)&(0.1, 0.3)/(0.4, 0.7)&(0.1, 0.2)/(0, 0.15)&(0.1, 0.15)/(0.1, 0.15)\\

\hline
H3&(70, 100)&(-2.5, 2)&(0.6, 0.8)/(0.75, 0.9)&(0.3, 0.8)/(0.6, 0.8)&(0.4, 0.6)/(0.45, 0.65)&(0.1, 0.25)/(0.3, 0.5)&(0.1, 0.15)/(0.1, 0.2)&(0.45, 0.65)/(0.45, 0.6)\\

\hline
H4&(160, 200)&(-6, 5)&(0.75, 0.8)/(0.75, 0.9)&(0.9, 1)/(0.8, 0.9)&(0.25, 0.6)/(0.4, 0.75)&(0.1, 0.25)/(0.2, 0.4)&(0.1, 0.15)/(0.3, 0.4)&(0.65, 0.8)/(0.45, 0.6)\\

\hline
\end{tabular}
}
}

    \end{center}
\end{table*}

\section{Further Experimental Results}
In this section, we supplement the implementation details of all experiments and provide additional experiments on the recovery network in CI\&Seg. The settings and aberrations distributions of VPL samples in BT are provided in Sec.~\ref{sec:teacher}. In Sec.~\ref{sec:ki}, the processing of KITTI-360 is depicted. We illustrate the training and evaluation settings of all related methods in Sec.~\ref{sec:details}. Finally, the analysis of computational overhead for applied solutions is depicted in Sec.~\ref{sec:qua}. 

\subsection{Datasets for Bidirectional Teacher}
\label{sec:teacher}
\begin{figure}[!t]
  \centering
  \includegraphics[width=1.0\linewidth]{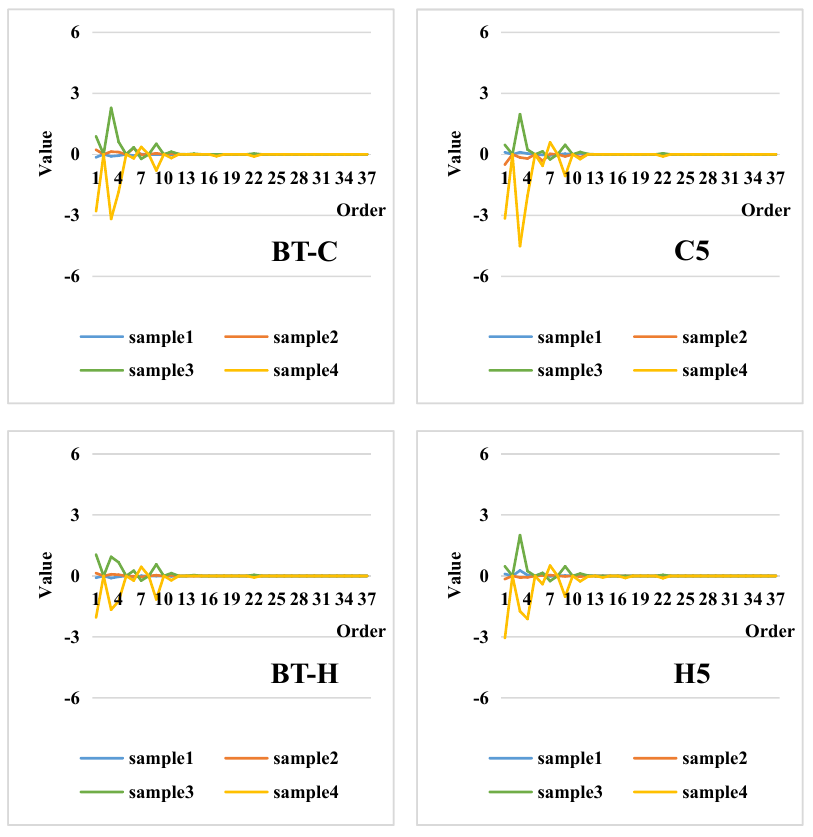}
  \caption{Average Zernike coefficient distributions over FoVs and RGB channels. We compare the selected 4 samples for BT training (BT-C and BT-H) with those for our benchmark (C5 and H5). The different aberration distributions simulate the synthetic-to-real gap. }
  \label{fig:btvpl}
\end{figure}

The dataset of BT represents the datasets of most computational imaging methods, which could be produced by ray-tracing~\cite{chen2021optical}, wave-based simulation~\cite{10021856}, or live-shooting~\cite{Peng2019LearnedLF}. In our case, we select wave-based simulation as the data generation method due to the established simulation model. Considering the potential synthetic-to-real gap where the simulated imaging results are often different from that of real MOS due to manufacture, the accurate PSFs of applied MOS are often unavailable for generating synthetic aberration images for CI training.
In other words, the calculated PSFs based on the design parameters of MOS often deviate from the real ones. In the case of our VPL samples, to simulate the gap, we should fine-tune the parameters and reconstruct other $20$ VPL samples different from those in our benchmark, for dataset generation of BT training.
TABLE~\ref{tab:btvpl} and Fig.~\ref{fig:btvpl} show the random ranges and average Zernike coefficient distributions of VPL samples for BT training. Compared to those in TABLE~\ref{tab:vpl} and Fig~\ref{fig:vplsample}, the fine-tuned parameters ensure that the aberration distribution deviates from the actual one of applied VPL samples, but with similar behavior and level for the effectiveness of stored prior CI knowledge. 

\begin{table}[!t]
    \begin{center}
        \caption{The fine-tuned ranges for generation of VPL samples used in BT training. The random ranges of the mentioned specific orders remain unchanged for similar aberration distribution.}
        \label{tab:btvpl}
        \resizebox{0.45\textwidth}{!}{
\setlength{\tabcolsep}{1mm}{    
\begin{tabular}{l|cc||c|cc}

\hline

&\textbf{Radius}&\textbf{Overall}&&\textbf{Radius}&\textbf{Overall}\\

\hline
BT-C1&(5, 45)&(-0.45, 0.35)&BT-H1&(20, 30)&(-0.4, 0.4)\\

\hline
BT-C2&(10, 75)&(-0.6, 1)&BT-H2&(25, 40)&(-0.7, 0.7)\\

\hline
BT-C3&(15, 155)&(-3.5, 2)&BT-H3&(85, 115)&(-3, 2.5)\\

\hline
BT-C4&(15, 255)&(-6, 6)&BT-H4&(150, 185)&(-6, 6)\\

\hline
\end{tabular}
}
}

    \end{center}
\end{table}

\begin{table*}[!t]
    \begin{center}
        \caption{Implementation details of evaluated segmenters. We detail the configurations which are available in mmsegmentation~\cite{mmseg2020}. }
        \label{tab:detailseg}
        \resizebox{1\textwidth}{!}{
\setlength{\tabcolsep}{1mm}{    
\begin{tabular}{l|ccccccc}

\hline

\textbf{Method}&\textbf{Backbone}&\textbf{Decoder}&\textbf{Crop size}&\textbf{Testing size (Cs.)}&\textbf{Testing size (KI.)}&\textbf{Iterations}&\textbf{Configuration}\\

\hline\hline
FCN~\cite{long2015fully}&ResNet101&FCNHead&$512\times1024$&$512\times1024$&$376\times1408$&$80K$&fcn\_r101-d8\_fp16\_512x1024\_80k\_cityscapes\\

\hline
PSPNet~\cite{zhao2017pyramid}&ResNet101&PSPHead&$512\times1024$&$512\times1024$&$376\times1408$&$80K$&pspnet\_r101-d8\_512x1024\_80k\_cityscapes\\

\hline
DeepLabV3+~\cite{chen2018encoder}&ResNet101&ASPPHead&$512\times1024$&$512\times1024$&$376\times1408$&$80K$&deeplabv3plus\_r101-d8\_512x1024\_80k\_cityscapes\\

\hline
SETR~\cite{zheng2021rethinking}&ViT&SETRUPHead&$768\times768$&$512\times1024$&$376\times1408$&$80K$&setr\_vit-large\_pup\_8x1\_768x768\_80k\_cityscapes\\

\hline
SegFormer~\cite{xie2021segformer}&MiT-b5&SegFormerHead&$1024\times1024$&$512\times1024$&$376\times1408$&$160K$&segformer.b5.1024x1024.city.160k\\

\hline
SegNeXt~\cite{guo2022segnext}&MSCAN&LightHamHead&$1024\times1024$&$512\times1024$&$376\times1408$&$160K$&segnext.large.1024x1024.city.160k\\

\hline
\end{tabular}
}
}

    \end{center}
\end{table*}

\subsection{Processing of KITTI-360}
\label{sec:ki}
\begin{figure}[!t]
  \centering
  \includegraphics[width=1.0\linewidth]{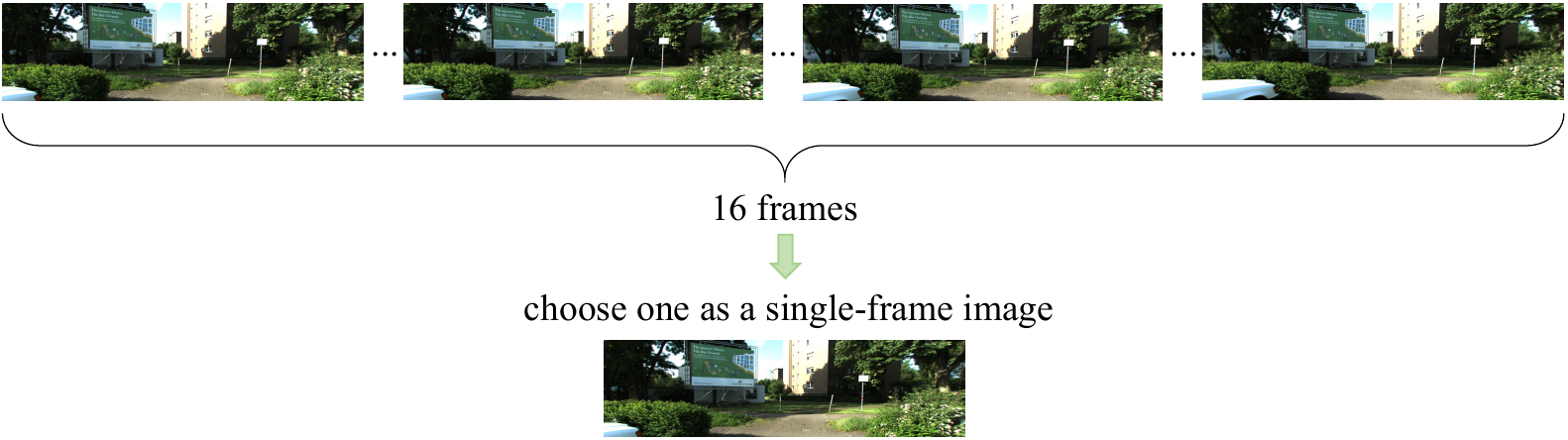}
  \caption{Illustration of our processing on KITTI-360. The image sequences are screened to be a single-frame image dataset. }
  \label{fig:kitti360_processing}
\end{figure}

In real-world scenarios, the corresponding clear image of the aberration image is not available. Consequently, the dataset of clear images must be different from that of aberration images. In other words, when the source clear images are from Cityscapes~\cite{cordts2016cityscapes}, the origin images for simulation should be another dataset with similar size, scenarios, and annotation rules. KITTI-360~\cite{liao2022kitti} is a good choice which is a dataset of image sequences with similar scenarios and annotation rules with Cityscapes. To ensure a similar size and data form of the two datasets, KITTI-360 is screened, where the frame rate is lowered by $16$ times. As is shown in Fig~\ref{fig:kitti360_processing}, the processed KITTI-360 is used as a single-frame dataset with diverse urban scenes. 

\subsection{Implementation Details}
\label{sec:details}
\PAR{Training of BT.}
We apply NAFNet~\cite{chen2022simple} for both the image recovery network and image degradation network in BT for its efficient and effective performance in low-level vision. Adapted from the attention block in Restormer~\cite{zamir2022restormer}, the backbone in NAFNet can deal with both spatially-variant and spatially-invariant blur with much fewer computational overheads. The trade-off between performance and efficiency makes NAFNet an optimal choice for BT.

The two networks in BT are trained separately for the recovery of target images and degradation of source images. The width and depth of applied NAFNet are $32$ and $[1,1,1,28]$. Based on BasicSR~\cite{basicsr}, in accordance with~\cite{chen2022simple}, we train BT with AdamW~\cite{loshchilov2017decoupled} for $160K$ iterations with an initial learning rate of $1e^{-3}$, which is gradually reduced to $1e^{-6}$ with the cosine annealing schedule~\cite{loshchilov2016sgdr}, on image patches of $256{\times}256$ with a batch size of $8$.

\PAR{Training of CIADA.}
We adopt MiT-b5~\cite{xie2021segformer} with a feature pyramid of $[64, 128, 320, 512]$ for the encoder, and Hamburger~\cite{guo2022segnext} which aggregates the last three stages of features with $C{=}512$ for both the semantic segmentation decoder and the auxiliary recovery decoder. Following~\cite{hoyer2022daformer} and~\cite{tranheden2021dacs}, we set $T{=}0.01$ for RCS temperature, $r{=}0.75$ and $\lambda_{FD}{=}0.005$ for FD, $\alpha{=}0.99$, and $\tau{=}0.968$ for data mixing. CIADA is trained with AdamW for $40K$ iterations with a learning rate of $6e^{-5}$ for the encoder and $6e^{-4}$ for the decoder, where weight decay of $0.01$ and a linear learning rate warmup of $1.5K$ iterations are applied. We train it with a batch size of $2$ on two $376{\times}376$ random crops based on mmsegmenmtation~\cite{mmseg2020}. 

\PAR{Evaluation of state-of-the-art segmenters.}
In TABLE~\ref{tab:detailseg}, we show the details of each evaluated segmenter. All settings and codes are available in mmsegmentation~\cite{mmseg2020} and~\cite{guo2022segnext}.
It is worth mentioning that the closest resolution to our tested images ($512{\times}1024$) within the released model weights of SETR is $768{\times}768$, which results in the worse performance of SETR in TABLE~\ref{tab:sota_seg} of the paper than that in~\cite{zheng2021rethinking}. Such a transformer-based method suffers from the weakness of being sensitive to the resolution of tested images~\cite{beyer2023flexivit,tian2023resformer}.

\PAR{Evaluation of possible solutions to SSOA.}
SegFormer-b5~\cite{xie2021segformer} is adopted for the Oracle and Src-Only methods. We train it with the same settings as our CIADA for a fair comparison.
Additionally, DAFormer~\cite{hoyer2022daformer} is chosen as the baseline of ST-UDA which is trained under its default setting except that the crop size is changed from $512{\times}512$ to $376{\times}376$ considering the resolution of KITTI-360~\cite{liao2022kitti} ($376{\times}1408$). 

\begin{table}[t!]
\begin{center}
\caption{Runtime analysis of representative segmentation models.}
\label{tab:sup1}
\resizebox{0.45\textwidth}{!}
{
\setlength{\tabcolsep}{2.5mm}{
\begin{tabular}{c|cccc}
\hline
\textbf{Method}     & DeepLabv3+ & SegFormer & SegNeXt & \textbf{Ours} \\ \hline
\textbf{Runtime (s)} &  0.354          & 0.336    &0.208         & 0.332         \\ \hline
\end{tabular}
}
}
\end{center}
\end{table}

\begin{table}[t!]
\begin{center}
\caption{Runtime analysis of representative recovery models in CI\&Seg.}
\label{tab:sup2}
\resizebox{0.45\textwidth}{!}
{
\setlength{\tabcolsep}{1mm}{
\begin{tabular}{c|ccc}
\hline
\textbf{Method}     & KPN   & Restormer & \textbf{NAFNet} \\ \hline
\textbf{Runtime (Rel.)} & 0.037s (11.14\%) & 0.190s (57.23\%)    & 0.068s (20.48\%) \\\hline
\end{tabular}
}
}
\end{center}
\end{table}

\subsection{Analysis of Computational Overhead}
\label{sec:qua}
The computational overhead is a key index of SSOA in real applications.
For the ST-UDA, the computational overhead of inference depends only on the applied segmentation model. 
We evaluate the runtime of our model and three representative models in TABLE~\ref{tab:sup1}.

In our framework, we consider the efficiency via applying the framework without additional computational overhead during inference compared to CI$\&$Seg.
The runtime of $3$ ablated recovery network is also shown in TABLE~\ref{tab:sup2}, where the relative runtime to our segmenter (\ie~Rel.) is provided in ``()''.
With $11.14\%{\sim}57.23\%$ more computational overhead, CI\&Seg brings limited improvements over Src-Only, where CIADA outperforms both of them by a large margin only with a segmenter during inference. 
Moreover, our CIADA is a plug-and-play framework for training segmenters of any architecture, which can benefit from the booming development of powerful and efficient segmentation model design.

\end{document}